\RequirePackage[2020-02-02]{latexrelease}
\documentclass[shortpaper]{clv3}
\usepackage{times}
\usepackage{latexsym}

\usepackage{graphicx} 
\usepackage{float}
\usepackage{lingmacros}
\usepackage{hyperref}
\usepackage{booktabs}
\usepackage{amsmath}
\usepackage{tablefootnote}
\usepackage{multirow}
\usepackage{appendix}
\usepackage{titlesec}

\newcommand{\appsection}[1]{
   \setcounter{table}{0}
   \setcounter{figure}{0}
   \setcounter{equation}{0}
  \section[\appendixname \thesection]{#1}%
}

\usepackage[utf8]{inputenc}
\usepackage{CJKutf8}
\usepackage[greek,russian,main=english]{babel}

\usepackage{xcolor,pifont}
\newcommand*\colourcheck[1]{%
  \expandafter\newcommand\csname #1check\endcsname{\textcolor{#1}{\ding{52}}}%
}
\newcommand*\colourcross[1]{%
  \expandafter\newcommand\csname #1cross\endcsname{\textcolor{#1}{\ding{56}}}%
}
\colourcheck{green}
\colourcross{red}

\usepackage{fancyhdr}
\newcommand{\mnlang}[1]{\texttt{[n.}\textbf{#1}\texttt{]}}
\newcommand{\mslang}[1]{\texttt{<n.}\textbf{#1}\texttt{>}}

\newcommand{\qgsample}[3]{
\begin{quote}
    \textbf{Sentence:} #1\\
    \textbf{Question:} #2\\
    \textbf{Suggested answer:} #3
\end{quote}
}

\issue{XX}{X}{XXXX}

\runningtitle{Quinductor}

\runningauthor{Dmytro Kalpakchi}

\begin{document}

\title{Quinductor: a multilingual data-driven method for generating reading-comprehension questions using Universal Dependencies}

\author{Dmytro Kalpakchi\thanks{E-mail: dmytroka@kth.se}}
\affil{Division of Speech, Music and Hearing, KTH Royal Institute of Technology, Stockholm, Sweden}

\author{Johan Boye}
\affil{Division of Speech, Music and Hearing, KTH Royal Institute of Technology, Stockholm, Sweden}

\maketitle

\begin{abstract}
We propose a multilingual data-driven method for generating reading comprehension questions using dependency trees. Our method provides a strong, mostly deterministic, and inexpensive-to-train baseline for less-resourced languages. While a language-specific corpus is still required, its size is nowhere near those required by modern neural question generation (QG) architectures. Our method surpasses QG baselines previously reported in the literature and shows a good performance in terms of human evaluation.
\end{abstract}

\section{Introduction}
\label{sec:intro}
We are interested in \textbf{question generation} (QG) -- the task of automatically generating reading comprehension questions and their correct answers from given declarative sentences. Numerous methods have been proposed for solving this task, most of which have been aimed at the English language. Recent methods are based on neural networks and rely on the availability of large-scale datasets, such as SQuAD \citep{rajpurkar2016squad} -- a question-answering dataset repurposed for QG -- or large-scale pretrained models, such as GPT-3 \citep{brown2020language}. Early methods, mostly based on context-free grammars, relied on the strict word order and the limited inflectional morphology of English. These traits made it relatively straightforward to craft hand-written templates based on these grammars. The above mentioned idiosyncracies and the unique availability of large-scale resources for English leave a number of open challenges for developing QG methods applicable to languages other than English.

The first challenge is the lack of large-scale training datasets, and a prohibitively high cost of obtaining such resources. State-of-the-art QG methods for English train their models on the previously mentioned SQuAD dataset, which contains more than 100,000 questions. Obtaining a good-quality dataset of a similar size is very expensive, especially for languages with fewer native speakers around the world.

The second challenge is knowing how well available methods developed for English would generalize to other languages, especially synthetic ones with richer inflectional morphology and less strict word order (e.g., Finnish, Turkish or Russian). To the best of our knowledge, not much research has been done on QG for these kinds of languages. 

The third challenge is assessing the obtained performance results. Evaluation results in isolation do not provide a comprehensive picture of the method's performance, especially when using only automatic evaluation metrics, such as BLEU \citep{papineni2002bleu}. Researchers that developed the first statistical QG methods for English could compare their results to baselines that relied on context-free grammars. However, most other languages lack QG baselines, leaving researchers to wonder if the obtained performance is worth the spent computational resources on training the model.

In this article we are addressing all three challenges by proposing a novel, mostly deterministic method, called \textbf{Quinductor}\footnote{The code is available at \url{https://github.com/dkalpakchi/quinductor}} (Question inductor), for automatically generating question-answer pairs from data. Quinductor is based on dependency trees and can also be used for languages other than English, due to the Universal Dependencies (UD) framework \citep{nivre2020universal} offering more than 200 treebanks in 100 languages. The method does require a language-specific QA dataset, but its size can be orders of magnitude smaller than SQuAD. Hence we believe that Quinductor can serve as a strong QG baseline for less-resourced languages.

\section{Related work}
\label{sec:related}
\citet{rus2008question} broadly defined QG as automatic generation of questions from inputs such as text, raw data or knowledge bases. In this article, we are interested in generating reading comprehension questions from textual data, with their respective correct answers, and we want to do this in multiple languages. We exclude Yes/No-questions and fill-in-the-blank questions, as those can be generated with less sophisticated methods \citep{gates2011generate,mostow2012generating,agarwal2011automatic}. Hence we limit the scope of related works only to articles exploring a similar QG setup.

To the best of our knowledge, no other work has proposed an automatic multilingual QG method relying on dependency parsing. The closest by spirit is the work by \citet{afzal2014automatic}, where sentences are matched against a set of automatically extracted semantic patterns from the GENIA Event Annotation corpus using a Named Entity Recognizer (NER). These patterns are used to extract relevant parts of the dependency tree, which are then transformed into the question by abstracting away the information constituting the correct answer (which should not be a part of the question). The method requires resources that are often lacking for other languages, such as, a NER system and a corpus which would be very expensive to annotate.

\citet{mazidi2015leveraging} also relied on a dependency parser, a semantic role labeler and discourse cues. However, their method only generated questions without the correct answers, requiring the manual creation of question templates, and relying on language-specific information. Similarly, \citet{khullar2018automatic} proposed a method using a dependency parser and three manually crafted rule sets for transforming statements into questions (without exploring the generation of correct answers).

Other non-neural QG methods utilised hand-written templates based on context-free grammars. One example is the work by \citet{heilman2009question}, which used an overgenerate-and-rank strategy for QG without generating correct answers. Another example is the work by \citet{bernhard2012question}, which is based on constituent trees and a NER system to generate questions (and their correct answers) in French. Such methods require linguists to create context-free grammars, which is an expensive process, especially for languages with less strict word order and a richer morphology than English.

The most recent QG methods are based on neural networks, and thus require both large-scale datasets in the language of interest, as well as vast computational resources to train the models. Impressive performance for English have been demonstrated by both Transformer-based masked language models \citep{chan2019recurrent,liao2020probabilistically,dong2019unified} and auto-regressive models based on encoder-decoder architectures \citep{kim2019improving,liu2019learning,du2017learning,song2018leveraging,zhao2018paragraph,bahuleyan2017variational}. Note that neural models typically do not generate correct answers, but instead use them as an input along with the sentence to generate questions. However, we are not going into more details on the neural methods, as our proposed method is not neural.

To the best of our knowledge, only a very limited number of neural methods explore QG in other languages than English, or multilingual QG. One such example is the work by \citet{kumar2019cross} exploring joint cross-lingual training aimed at reusing the large-scale SQuAD dataset for Hindi and Chinese.

\section{Methodology}
Let $D$ be a dataset consisting of triples $(c_i, q_i, a_i)$, where $c_i$ is a context (a text passage), $q_i$ is a question created based on $c_i$, and $a_i$ is a contiguous phrase in~$c_i$, answering $q_i$. A pair of $(q_i, a_i)$ will be referred to as a question-answer pair (QA-pair). The aim is then to be able to generate QA-pairs $(q_j', a_j')$ given a previously unseen context $c_j'$.

\begin{figure}[t]
	\centering
	\includegraphics[width=\textwidth]{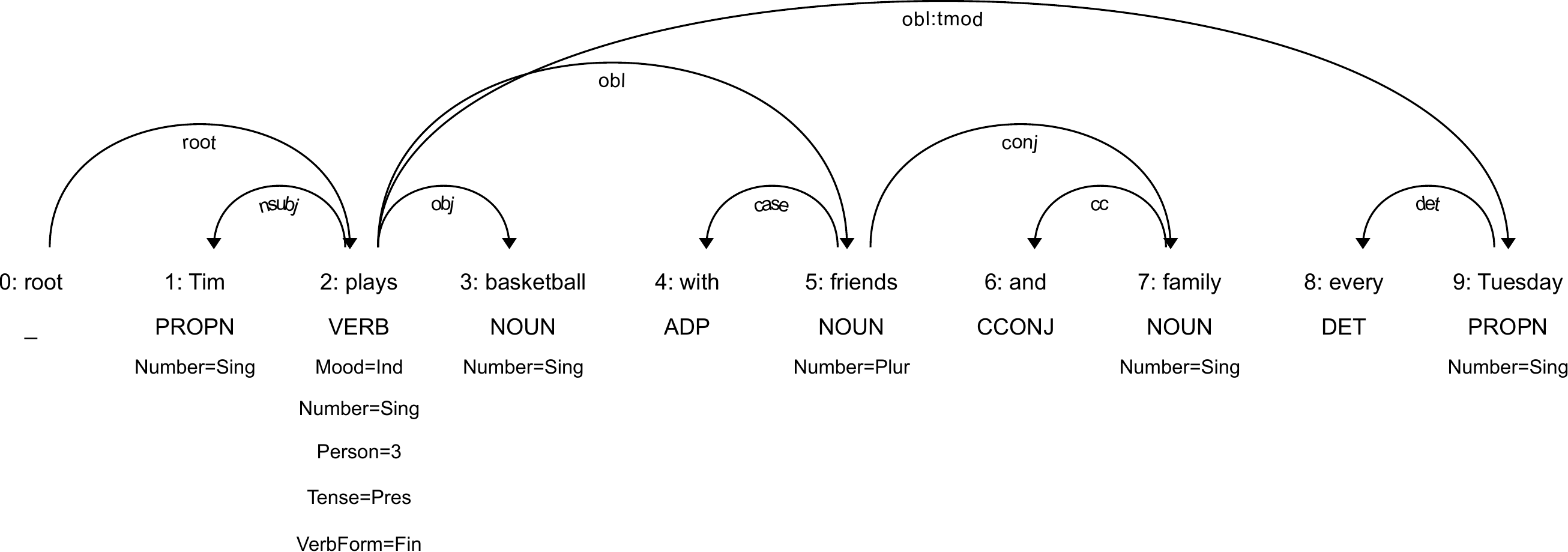}
	\caption{The dependency tree for the sentence ``Tim plays basketball with friends and family every Tuesday''}
	\label{fig:en_dep_example}
\end{figure}

Our method, Quinductor, automatically induces QA-templates from the dataset~$D$ using dependency parsing based on the UD framework. More formally, let $s_i$ be the sentence from the context $c_i$ in which the answer $a_i$ appears ($s_i$~can be found using a sentence segmenter, a tokenizer, and simple string matching). The QA-pair $(q_i, a_i)$ is recast into a template in a specific formal language (defined in Section \ref{sec:temp_language}), using parts of the dependency tree for the sentence~$s_i$. For instance, suppose $s_i$ is \emph{``Tim plays basketball with friends and family every Tuesday''} (with its dependency tree shown in Figure \ref{fig:en_dep_example}), and $q_i$ is ``When does Tim play basketball with friends and family?''. Assuming $r$ represents the root of the dependency tree (i.e.\ the dependent of the ``root'' pseudonode; the word ``plays'' in this example), $q_i$ can now be expressed using the question template (\ref{ex1_q}) and the answer ``every Tuesday'' could be expressed using the answer template (\ref{ex1_a}). For a formal definition of these expressions we refer to the Section \ref{sec:temp_language}.
\enumsentence{\texttt{When does [r.nsubj\#1] [r.lemma] [r.obj\#3] <r.obl\#5>?}\label{ex1_q}}
\enumsentence{\texttt{<r.obl:tmod\#9>}\label{ex1_a}}

Such a transformation can be applied only if certain conditions are met, and therefore each QA-template has an associated {\bf guard\/} (described in its own formal language defined in section \ref{sec:guard_language}). After inducing both QA-templates and associated guards, we can then apply them to any previously unseen context $c'$ by processing its every sentence $s'$ using the following procedure.
\begin{algorithm}
    \item[Step 1] Perform dependency parsing on $s'$ and get a dependency tree $T'$.
    \item[Step 2] Find all satisfied guards for $T'$ and get a set of corresponding QA-templates $QA_{T'}$.
    \item[Step 3] Apply all templates from $QA_{T'}$ to $s'$, in order to get a set of generated question-answer pairs $QA'$. Note that many QA-pairs will be unsatisfactory, which is why the next step is introduced.
    \item[Step 4] Rank $QA'$ so that a QA-pair $(q',a')$ is ranked highly if it is likely to be relevant, grammatical, and where $a'$ is likely to be the correct answer to $q'$. The ranking is done according to the method presented in Section \ref{sec:ranking}.
\end{algorithm}

\begin{figure}[b]
	\centering
	\includegraphics[width=\textwidth]{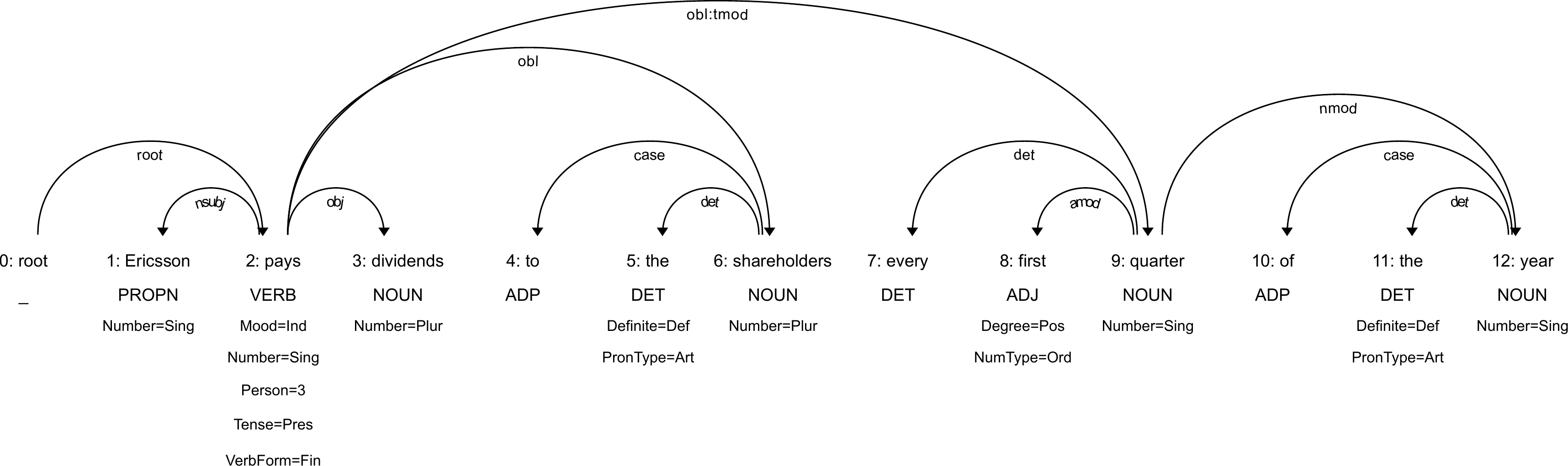}
	\caption{The dependency tree for the sentence ``Ericsson pays dividends to the shareholders every first quarter of the year''}
	\label{fig:en_dep_example2}
\end{figure}
As an example of this procedure, consider
$s'$ to be the sentence ``Ericsson pays dividends to the shareholders every first quarter of the year'' (with a dependency tree in Figure \ref{fig:en_dep_example2}), then using the question template (\ref{ex1_q}) and the answer template (\ref{ex1_a}), the following QA-pair can be generated:
\enumsentence{When does Ericsson pay dividends to the shareholders? -- Every first quarter of the year}

The key to our method is to make templates as generic as possible to allow a certain amount of variation in their dependency structures. For instance, we want to avoid using adverbial clauses word by word, but instead matching adverbial clauses more generally. This generalization is addressed by \textbf{a novel shift-reduce algorithm}, described in Section \ref{sec:temp_induction}. Automatic induction of guards is described in \ref{sec:guard_induction}. However, before describing the induction algorithms, let us first explain and motivate the designed template and guard languages. In the sections below we use a bold font in the expression definitions to indicate metalinguistic variables which are not part of the defined languages.

\subsection{Template language}
\label{sec:temp_language}
Let $T$ be an arbitrary dependency tree and \texttt{r} denote \emph{the root of the dependency tree}, i.e. the dependent of the ``root'' pseudonode of $T$. In the example sentence in Figure \ref{fig:en_dep_example} \texttt{r} corresponds to the word ``play'' (all further examples will also be given for this sentence). Let \texttt{n} be an arbitrary node of $T$, then the following definitions are introduced.
\begin{itemize}
    \item \texttt{n.}\textbf{rel\#id} denotes a dependent of \texttt{n} with a dependency relation \textbf{rel} and index \textbf{id} of this dependent of \texttt{n} (starting from 0 for the ``root'' pseudonode). For instance, \texttt{r.obj\#3} denotes the node for the word ``basketball''.
    \item \texttt{n.}\textbf{rel1\#id1}.\textbf{rel2\#id2} ... .\textbf{relN\#idN} denotes a node \texttt{n'} such that there exists a directed path between nodes \texttt{n} and \texttt{n'} with each edge having a corresponding dependency relation from a relation chain \textbf{rel1\#id1}.\textbf{rel2\#id2} ... .\textbf{relN\#idN}. The node \texttt{n'} will be referred to as \emph{a node at the end of the chain} and a whole relation chain will be shortened to \textbf{relchain}. For instance, \texttt{r.obl\#5.case\#4} denotes the node for the word ``with''. This node can then be referred as the node at the end of the chain \texttt{obl\#5.case\#4}. The ID of the last element of relchain is later referred to as the ID of a template expression.
\end{itemize}

The IDs are included in the template expressions above to be able to distinguish between different dependents having the same dependency relation. To illustrate when this could be necessary, imagine the dependency relation between the words ``plays'' and ``Tuesday'' is {\tt obl} instead of {\tt obl:tmod} (an inaccuracy that could be produced by the dependency parsers in practice, especially for languages other than English). Then the question ``When does Tim play basketball with friends and family?'' with the answer ``every Tuesday'' would result into the following QA-template (assuming IDs are excluded):
\enumsentence{\texttt{When does [r.nsubj] [w.lemma] <r.obl>?}\label{ex1a_q}}
\enumsentence{\texttt{<r.obl>}\label{ex1a_a}}
It is impossible to distinguish \texttt{<r.obl>} in the question template (\ref{ex1a_q}) from the one in the answer template (\ref{ex1a_a}). However if the IDs are introduced, then one immediately understands that those expressions correspond to different nodes. Note that differentiating between nodes with the same dependency relations is the only purpose of IDs, i.e. we do NOT require the new sentences using QA-templates to have exactly the same IDs, as it would obviously hinder generalization.

Let us define the following operators for selecting substructures from a dependency tree $T$:
\begin{itemize}
    \item \texttt{[n]} extracts the token at the node \texttt{n} (for instance, \texttt{[r]} extracts the token ``plays'');
    \item \mnlang{relchain1} extracts the token of the node at the end of \textbf{relchain1} (for instance, \texttt{[r.obl\#5.conj\#7]} extracts the token ``family'');
    \item \texttt{[n.lemma]} (\texttt{[n.}\textbf{relchain}\texttt{.lemma]}) extracts the lemma of the token at the node {\tt n} (the node at the end of \textbf{relchain1}). Either of these will be referred to as a \textbf{lemma-expression}. For instance, \texttt{[r.lemma]} extracts the string ``play'' and \texttt{[r.obl\#5.lemma]} extracts ``friend''.
    \item \texttt{<n>} extracts the text string of the subtree rooted at the node \texttt{n} (for instance, \texttt{<r>} extracts the whole sentence);
    \item \mslang{relchain1} extracts the text string of the subtree rooted at the node at the end of \textbf{relchain1} \emph{preserving the linear order} (for instance, \texttt{<r.obl\#5.conj\#7>} extracts the string ``and family'');
    \item \mslang{relchain1 - relchain2} extracts the text string of the subtree rooted at the node at the end of \textbf{relchain1} \emph{except the text string of the subtree rooted at the node at the end of} \textbf{relchain2} (if such a subtree exists). Relchains that are subtracted in any template expression will be referred to as \textbf{negatives}. For instance, \texttt{<r.obl\#5.conj\#7 - cc\#6>} extracts the string ``family''. However, non-existing negatives do not influence the result, hence \texttt{<r.obl\#5.conj\#7 - case\#6>} extracts the string ``and family'', since there is no child of \texttt{r.obl\#5.conj\#7} with a dependency relation \texttt{case}.
    \item \mslang{relchain1 - relchain2*} extracts the text string of a subtree rooted at the node at the end of \textbf{relchain1} \emph{except the contents of the node at the end of} \textbf{relchain2} (if it exists). For instance, \texttt{<r.obl\#5 - conj\#7}*\texttt{>} extracts the string ``with friends and''. Note that the extracted string is not guaranteed to be a contiguous substring of the sentence.
\end{itemize}
Template expressions surrounded by square brackets (\texttt{[]}) will be referred to as \textbf{node-level expressions}, and those surrounded by angle brackets (\texttt{<>}) as \textbf{subtree-level expressions}.


To distinguish the answer from the question, we use an additional binary infix operator\break\texttt{q => a}, denoting that the first operand is the question template, and the second one is the answer template. For instance, the question ``When does Tim play basketball with friends and family?'' with the answer ``every Tuesday'' could be represented as the following QA-template.
\enumsentence{\texttt{When does [r.nsubj\#1] [r.lemma] [r.obj\#3] <r.obl\#5>? => <r.obl:tmod\#9>}\label{sent:tquest}}
Note that words from the question that do not appear in the original sentence will not form any template expressions. Instead, they will be considered \textbf{constant} and rendered as a plain text, e.g., ``When'' and ``does'' in the template (\ref{sent:tquest}).


\subsection{Guard language}
\label{sec:guard_language}
Recall that a {\bf guard\/} is an expression specifying conditions for using a specific template. Formally, let $T$ be an arbitrary dependency tree, and \texttt{n} denote a node in $T$. Then:
\begin{itemize}
    \item \texttt{n.pos} denotes the part-of-speech (POS) tag assigned to the word associated with~\texttt{n} (below referred to as a \textbf{pos-property});
    \item \texttt{n}\texttt{.morph} denotes a set of morphological features, as defined by UD, associated with \texttt{n} (below referred to as a \textbf{morph-property}).
\end{itemize}
Each guard consists of clauses separated by a comma operator (\texttt{,}) denoting logical AND. Let us introduce operators defining the conditions for the guard clause to be satisfied:
\begin{itemize}
    \item the unary operator \texttt{exists} can be applied exclusively to relchains in order to only accept sentences having a specified relchain;
    \item a binary operator \texttt{is} (\texttt{is\_not}) can be applied merely to pos-properties to only accept sentences with a specific node having (lacking) a specified POS-tag;
    \item a binary operator \texttt{has} (\texttt{has\_not}) can be applied to morph-properties exclusively to only accept sentences with a specific node having (lacking) specified morphological properties (in the UD format).
\end{itemize}

To specify which template should be used if all guard clauses are satisfied, we use an additional infix operator \texttt{guard -> t} denoting that if the first operand (guard) is satisfied, the template found by the unique identifier \texttt{t} can be used.

To exemplify, the guard for the template (\ref{sent:tquest}) could look as follows.
\enumsentence{\texttt{n.pos is VERB, n.nsubj exists, n.obj exists, n.obl exists,\\ n.obl:tmod exists -> template3}}
Note that \emph{no requirement} on the exact IDs of the nodes is present in the guards, since the IDs are only used during the template induction phase.

After having described both template and guard languages, we are now ready to explain the algorithms for automatically inducing templates (Section \ref{sec:temp_induction}) and guards (Section \ref{sec:guard_induction}).

\subsection{Template induction}
\label{sec:temp_induction}
Recall that a datapoint is a triple $(c_i, q_i, a_i)$, where $c_i$ is a context, $q_i$ is a question asked on the basis of $c_i$, and $a_i$ is a contiguous phrase in $c_i$ constituting the correct answer to $q_i$. The goal is to induce templates for every $(q_i, a_i) \in D$, allowing to generalize to syntactically similar QA-pairs $(q'_j, a'_j) \not\in D$. This is achieved by merging template expressions into subtree-level expressions as much as possible, using the novel shift-reduce algorithm described below.

The preprocessing step is to find all triples $(s_i, q_i, a_i)$ such that $a_i$ is a contiguous phrase in $s_i \in S(c_i)$, where $S(c_i)$ is the set of sentences of the context $c_i$. Recall that this step is trivially performed using a sentence segmenter, a tokenizer, and simple string matching.

The next step is to select only \textbf{satisfactory triples}, where $s_i$ and $q_i$ have at least one word in common (if not, then generalization is impossible). After obtaining a number of satisfactory triples $(s_i, q_i, a_i)$, the induction of a template for transforming $s_i$ into a pair of $(q_i, a_i)$ can be described as the following 3-step process applied twice (once for $q_i$ and once for $a_i$).
\begin{enumerate}
    \item \textbf{Sentence transformation}. Describe every word of $q_i$ ($a_i$) in terms of dependency structures present in $s_i$ using the formal template language presented in Section \ref{sec:temp_language}. When finished, proceed to step 2.
    \item \textbf{Shift-reduce}. Simplify the template obtained at the previous step using the novel shift-reduce algorithm described in Section \ref{sec:shift_reduce}. In the rare case when the resulting template consists only of a single template expression (and would therefore generalize poorly), return the sentence transformation from step~1 as the final template, otherwise, proceed to step~3.
    \item \textbf{Merging negatives}. If possible, merge negatives (the subtracted relchains) in every template expression using the algorithm described in Section \ref{sec:neg_collapse}. Return the template with merged negatives as final.
\end{enumerate}

Recall that words from the question that do not appear in the original sentence will be considered \textbf{constant} and rendered as a plain text. Templates containing only constants will not generalize and are thus removed after all templates have been induced. The remaining templates (i.e., with at least one non-constant template expression) are post-filtered to exclude templates with rare words (since those will not generalize well). We define a word as rare if it appeared in less than 25\% of the documents and detect it based on the inverse document frequency (IDF), i.e. we exclude all templates with a maximal IDF among their constants exceeding $log(\frac{N}{\frac{N}{4}})=log(4)$, where $N$ is the number of documents in the corpus.

\subsubsection{Sentence transformation}
\label{sec:s_transform}
The goal of this step is to describe every word in $q_i$ and $a_i$ in terms of dependency structures of $s_i$. For instance, consider the QA-pair ``When does Tim play basketball with friends? -- Every Tuesday'', created based on the example sentence (see Figure \ref{fig:en_dep_example}). Sentence transformation applied to the question would then look as follows:
\enumsentence{\texttt{When does [r.nsubj\#1] [r.lemma] [r.obj\#3] [r.obl\#5.case\#4] [r.obl\#5]}\label{sent:s_trans_q}}
Whereas sentence transformation applied to the answer would take the following form:
\enumsentence{\texttt{[r.obl:tmod\#9.det\#8] [r.obl:tmod\#9]}\label{sent:s_trans_a}}

To perform sentence transformation, first, both $s_i$ and $q_i$ should be parsed to get the dependency trees $T_{s_i}$ and $T_{q_i}$ respectively. $T_{q_i}$ is then traversed in linear order $L_{T_{q_i}}$, skipping the question word, which is assumed to be the first word from the beginning or the end of the sentence, depending on the language of interest. For each node $n_q$ in $L_{T_{q_i}}$, the algorithm attempts to find a matching node in $T_{s_i}$ with the same token. If no matching nodes are found, $n_q$ is replaced by its token. If matching nodes are found, $n_q$ is replaced by the list of template expressions corresponding to those nodes in $T_{s_i}$ (using the template language presented in Section \ref{sec:temp_language}). This list is sorted in the ascending order by the distance from the root node of $s_i$ in edges. The resulting list of lists of template expression will be referred to as \textbf{LLTE}. Note that generation of lemma-expressions (e.g., {\tt [r.lemma]}) is subject to the availability of a lemmatizer, in the absence of which the algorithm will simply insert a constant expression (i.e., the token itself).

A template will generalize if many syntactic structures can be merged into subtree-level expressions, which is the goal of the shift-reduce step of Quinductor. Hence, the result of sentence transformation should contain as many long contiguous phrases as possible. With this goal in mind, after all nodes in $L_{T_{q_i}}$ have been processed, the combination of template expressions with longest contiguous spans is selected. This can be achieved by finding template(s) with the smallest sum of absolute ID differences between every two neighboring template expressions. 

\begin{figure}[t]
	\centering
	\includegraphics[width=\textwidth]{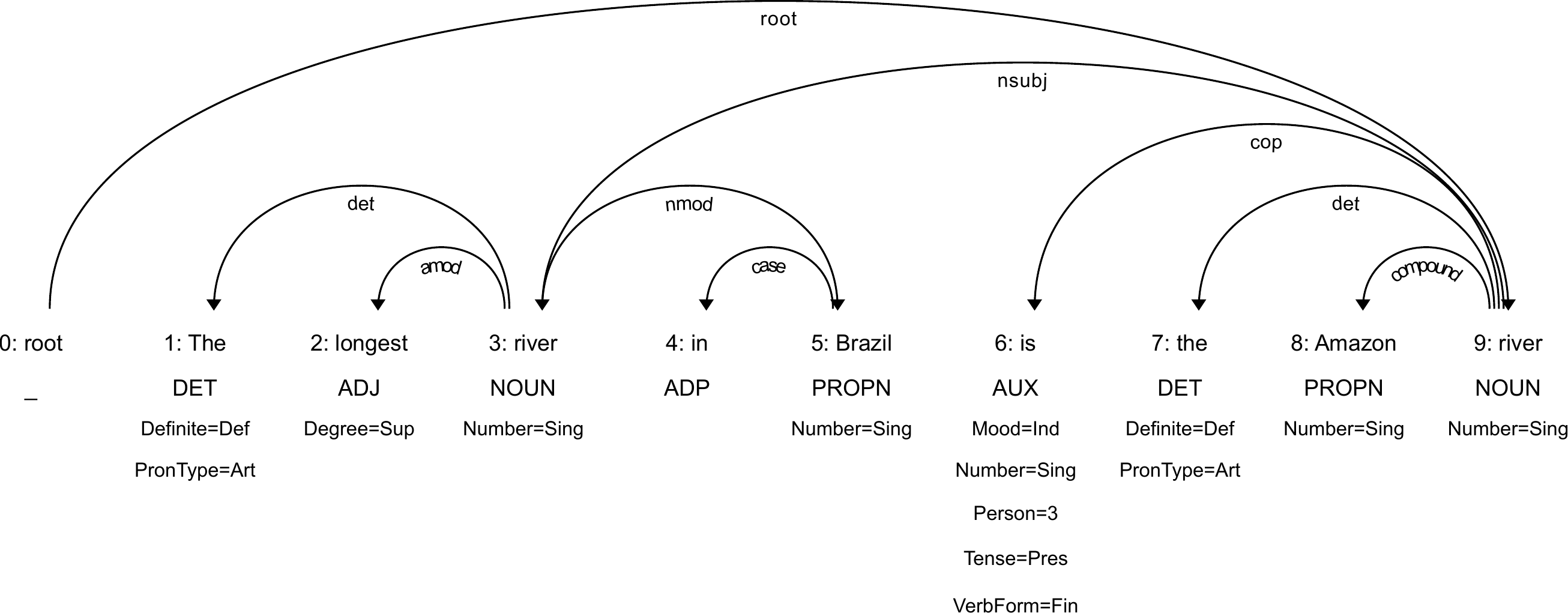}
	\caption{The dependency tree for the sentence ``The longest river in Brazil is the Amazon river''}
	\label{fig:en_dep_example3}
\end{figure}

For instance, consider the sentence ``The longest river in Brazil is the Amazon river'' with a dependency tree shown in Figure \ref{fig:en_dep_example3}. Assume that the question in the dataset which is based on this sentence  is ``What is the longest river in Brazil?''. The LLTE for this question is shown in Figure \ref{fig:contig_example}, assuming {\tt w} refers to the root of the original sentence, namely the word ``river''.
\begin{figure}[H]
    \centering
    \begin{tabular}{|p{0.7cm}p{1.3cm}p{1.3cm}p{1.8cm}p{0.6cm}p{2.5cm}p{2.2cm}|}
        \hline
        What & is & the & longest & river & in & Brazil\\
        \hline
        What & {\tt w.co\#6} & {\tt w.de\#7} & {\tt w.ns.am\#2} & {\tt w\#9} & {\tt w.ns.nm.ca\#4} & {\tt w.ns.nm\#5}\\
        & & {\tt w.ns.de\#1} & & {\tt w.ns\#3} & & \\
        \hline
    \end{tabular}
    \caption{LLTE for the question ``What is the longest river in Brazil?''. Each column represents all available alternatives for the given word. For the sake of brevity, the {\tt [$\cdot$]} operator is omitted, since all template expressions are node-level, the first two letters of dependency relations are used, and only the IDs of the last dependency relations in relchains are specified.}
    \label{fig:contig_example}
\end{figure}
As we can see, only the words ``the'' and ``river'' have multiple possible representations (i.e., {\tt w.det\#7} and {\tt w.nsubj\#3.det\#1} for ``the'' and {\tt w\#9} and {\tt w.nsubj\#3} for ``river''). The sums of absolute ID differences for different combinations of representations for the words ``the'' and ``river'' are presented in Table \ref{tab:contig_sums}.
\begin{table}[b]
\caption{\label{tab:contig_sums} Sums of absolute ID differences for alternative representations of the words ``the'' and ``river'' for LLTE in Figure \ref{fig:contig_example}}
\begin{tabular}{llc}
\midrule
\textbf{Representation of ``the''} & \textbf{Representation of ``river''} & \textbf{Sum of absolute ID differences} \\
\midrule
{\tt [w.det\#7]} & {\tt [w\#9]} & 19\\
{\tt [w.det\#7]} & {\tt [w.nsubj\#3]} & 9\\
{\tt [w.nsubj\#3.det\#1]} & {\tt [w\#9]} & 19\\
{\tt [w.nsubj\#3.det\#1]} & {\tt [w.nsubj\#3]} & 9\\
\bottomrule
\end{tabular}
\end{table}
The first representation from LLTE with the minimal sum of absolute ID differences is {\tt [w.det\#7]} for ``the'' and  {\tt [w.nsubj\#3]} for ``river'', resulting in the following sentence transformation
\enumsentence{What {\tt [w.cop\#6]} {\tt [w.det\#7]} {\tt [w.nsubj\#3.amod\#2]} {\tt [w.nsubj\#3]} {\tt [w.nsubj\#3.nmod\#5.case\#4]} {\tt [w.nsubj\#3.nmod\#5]}}
As can be seen, the algorithm chose the right expression for the word ``river'' and the wrong one for the word ``the''. Such errors depend on the order of lists in LLTE and there's no universal order that will result in choosing the right expressions all the time for all the languages.

\subsubsection{Shift-reduce}
\label{sec:shift_reduce}
The goal of this step is to make templates generalizable, which is achieved by merging template expressions into subtree-level expressions as much as possible using a novel shift-reduce algorithm. At every algorithm step, a current template (starting with the template obtained after the sentence transformation) is divided into \emph{a LIFO stack}, where all seen items reside, and \emph{a FIFO buffer}, containing the remainder. Depending on the stack-buffer configuration, one of the following two actions can be chosen:
\begin{itemize}
    \item \textbf{SHIFT}, that removes the top expression from the buffer and adds it to the stack.
    \item \textbf{REDUCE}, that checks the topmost and the second topmost expressions on the stack and merges them into a subtree-level expression.
\end{itemize}

While SHIFT action is self-explanatory, REDUCE can be described as a 3-step procedure, operating on the topmost ({\tt stackTop}) and the second topmost ({\tt stackTop2}) template expressions on the stack.

\begin{algorithm}
    \item[Step 1] Extract relchains from {\tt stackTop} and {\tt stackTop2} and then find the longest common prefix for them, later referred to as the \textbf{common relchain}. The node at the end of the common relchain is the closest common ancestor of the nodes corresponding to {\tt stackTop} and {\tt stackTop2}.
    \item[Step 2] The second step is to ensure that the two \textbf{merging conditions} are satisfied:
    \begin{enumerate}
        \item the common relchain is not empty;
        \item the common relchain differs from either relchain of {\tt stackTop} or {\tt stackTop2} by at most one dependency relation.
    \end{enumerate}
    We have empirically found that these merging conditions increase chances of generalization.
    \item[Step 3] If the aforementioned conditions are met, {\tt stackTop} and {\tt stackTop2} can be replaced by the template expression of their common relchain with a number of subtracted negatives corresponding to all tokens of the induced subtree except those necessary to keep: {\tt stackTop}, {\tt stackTop2}, any node from the sentence transformation, and any whole subtree containing any of these nodes.
\end{algorithm}

To illustrate the algorithm, consider turning the phrase ``friends and family'' from the sentence in Figure \ref{fig:en_dep_example} into a template. Initially the stack is empty and the buffer contains the sentence transformation of the phrase, resulting in the configuration shown in Figure \ref{fig:step1}.
\begin{figure}[H]
    \begin{tabular}{|p{0.46\textwidth}|}
        \hline
        \multicolumn{1}{|c|}{\textbf{Stack}}\\
        \hline\\
        \rule{0pt}{20.5pt}\\
        \hline
    \end{tabular}
    \begin{tabular}{|p{0.46\textwidth}|}
        \hline
        \multicolumn{1}{|c|}{\textbf{Buffer}}\\
        \hline
        {\tt [r.obl\#5] [r.obl.\#5.conj\#7.cc\#6] [r.obl\#5.conj\#7]}\\
        \hline
    \end{tabular}
    \caption{Initial stack-buffer configuration for the shift-reduce algorithm applied on the sentence transformation of the phrase ``friends and family'' from the sentence in Figure \ref{fig:en_dep_example}.}
    \label{fig:step1}
\end{figure}
First, two SHIFTs are required to ensure that the stack has at least two expressions, leading to the configuration in Figure \ref{fig:step2}.
\begin{figure}[H]
    \begin{tabular}{|p{0.46\textwidth}|}
        \hline
        \multicolumn{1}{|c|}{\textbf{Stack}}\\
        \hline
        {\tt [r.obl.\#5.conj\#7.cc\#6] [r.obl\#5]}\\
        \hline
    \end{tabular}
    \begin{tabular}{|p{0.46\textwidth}|}
        \hline
        \multicolumn{1}{|c|}{\textbf{Buffer}}\\
        \hline
        {\tt [r.obl\#5.conj\#7]}\\
        \rule{0pt}{8.5pt}\\
        \hline
    \end{tabular}
    \caption{Stack-buffer configuration after 2 SHIFT actions.}
    \label{fig:step2}
\end{figure}
The top two template expressions on the stack are neither constants nor lemma-expressions, so the REDUCE action can be invoked. The common prefix for relchains is {\tt obl\#5}, meaning both merging conditions are satisfied and the expressions can be merged into {\tt <r.obl\#5 - conj\#7}*{\tt>}. This new expression replaces the top two expressions on the stack, resulting in the configuration in Figure \ref{fig:step3}.
\begin{figure}[H]
    \begin{tabular}{|p{0.46\textwidth}|}
        \hline
        \multicolumn{1}{|c|}{\textbf{Stack}}\\
        \hline
        {\tt <r.obl\#5 - conj\#7}*{\tt>}\\
        \hline
    \end{tabular}
    \begin{tabular}{|p{0.46\textwidth}|}
        \hline
        \multicolumn{1}{|c|}{\textbf{Buffer}}\\
        \hline
        {\tt [r.obl\#5.conj\#7]}\\
        \hline
    \end{tabular}
    \caption{Stack-buffer configuration after 2 SHIFT and 1 REDUCE actions}
    \label{fig:step3}
\end{figure}
The next step is SHIFTing the last expression of the buffer into the stack, resulting in the configuration in Figure \ref{fig:step4}.
\begin{figure}[H]
    \begin{tabular}{|p{0.46\textwidth}|}
        \hline
        \multicolumn{1}{|c|}{\textbf{Stack}}\\
        \hline
        {\tt [r.obl\#5.conj\#7]}\\{\tt <r.obl\#5 - conj\#7}*{\tt>}\\
        \hline
    \end{tabular}
    \begin{tabular}{|p{0.46\textwidth}|}
        \hline
        \multicolumn{1}{|c|}{\textbf{Buffer}}\\
        \hline\\
        \rule{0pt}{8.5pt}\\
        \hline
    \end{tabular}
    \caption{Stack-buffer configuration after 3 SHIFT and 2 REDUCE actions}
    \label{fig:step4}
\end{figure}
The top two template expressions on the stack are neither constants nor lemma-expressions, so the REDUCE action can be invoked again. Following the same logic as before, the expressions can be merged into {\tt <r.obl\#5>}, resulting in the configuration in Figure \ref{fig:step5}.
\begin{figure}[H]
    \begin{tabular}{|p{0.46\textwidth}|}
        \hline
        \multicolumn{1}{|c|}{\textbf{Stack}}\\
        \hline
        {\tt <r.obl\#5>}\\
        \hline
    \end{tabular}
    \begin{tabular}{|p{0.46\textwidth}|}
        \hline
        \multicolumn{1}{|c|}{\textbf{Buffer}}\\
        \hline\\
        \hline
    \end{tabular}
    \caption{Stack-buffer configuration after 3 SHIFTs and REDUCE}
    \label{fig:step5}
\end{figure}
The buffer is empty, which means shift-reduce is finished. The final template is on the stack.

\subsubsection{Merging negatives}
\label{sec:neg_collapse}
Recall that negatives are relchains subtracted in any template expression. The goal of this step is to merge negatives in the resulting template after the shift-reduce step, in order to make templates even more generic and generalizable. For instance, sentence transformation (\ref{sent:s_trans_q}) would be converted to the following template after shift-reduce:
\enumsentence{\texttt{When does [r.nsubj\#1] [r.lemma] [r.obj\#3] <r.obl\#5 - conj\#7.cc\#6 - conj\#7}*\texttt{>}\label{sent:res_temp}}

The downside of this template is that it presupposes that by subtracting {\tt conj\#7.cc\#6} and {\tt conj\#7}* it effectively removes the whole subtree corresponding to {\tt conj\#7}. However, conjuncts vary in structure and thus this template will generalize poorly to sentences with syntactically similar structures. To avoid this, the negatives of each template expression should be merged as much as possible to their common parent.

To perform this step, create a mapping between each node and its direct children. Then, for every template expression, check if any subset of negatives matches any set of children from the mapping. In case of a match, swap all matched negatives for the corresponding subtree root. After these steps expression (\ref{sent:res_temp}) transforms into:
\enumsentence{\texttt{When does [r.nsubj\#1] [r.lemma] [r.obj\#3] <r.obl\#5 - conj\#7>}\label{sent:res_final}}

\subsection{Guard induction}
\label{sec:guard_induction}
A template can have multiple guards. Consider two sentences ``John is playing basketball'' and ``John has played basketball''. Both questions ``What is John playing?'' and ``What has John played?'' would result in the same template (supported by the previously mentioned sentences). However, the morphological properties of the root of each sentence (``playing'' and ``played'' respectively) are different. Hence, there are 2 different cases when this template could be applied, and thus 2 guards.

Motivated by the example above, guards consist of a base guard and complementary guards. \textbf{A base guard} contains the requirements for using templates and its creation involves the following 3 steps:
\begin{enumerate}
    \item Create an {\tt exists}-clause for the relchain of every template expression present in the question or answer (excluding the negatives).
    \item If a template for the answer contains a nominal subject ({\tt nsubj}) as a non-negative expression, add the clause {\tt n.nsubj.morph has\_not PronType=Rel} ensuring that the subject is not a relative pronoun (e.g., ``which''). This is motivated by the fact that no reading comprehension question would ask about a relative pronoun.
    \item If a root is involved in the creation of a template, and it is a verb without an auxiliary verb, add a clause {\tt n.aux not\_exists}. The rationale behind this step is to separate templates for questions with either copula or tenses requiring a modal verb, from those questions that do not exhibit these features. 
\end{enumerate}
\textbf{Complementary guards} contain requirements specific to the sentences supporting the generated template. Complementary guards are induced by creating an {\tt is}-clause for the pos-property and {\tt has}-clause for the morph-property of the root of every sentence from the corpus supporting the current template.

To get a final set of guards for the template, add the base guard to each complementary guard and use an infix operator {\tt ->} to point each guard from the induced set to the template of interest.

For instance, the guard for the template (\ref{sent:res_final}) would look as (\ref{sent:guard_final}), where {\tt props} equals to {\tt Mood=Ind|Number=Sing|Person=3|Tense=Pres|VerbForm=Fin}, and {\tt n.morph has props, n.pos is VERB} is the only auxiliary guard.
\enumsentence{\texttt{n.morph has props, n.pos is VERB, n.nsubj exists, n.obj exists, n.obl exists, n.nsubj.morph has\_not PronType=Rel, n.aux not\_exists -> template7}\label{sent:guard_final}}

\subsection{Ranking and filtering}
\label{sec:ranking}
After all templates and guards have been generated, they can be applied to unseen data to produce a number of QA-pairs. With the purpose of down-voting undesirable QA-pairs we use the following two models, that serve as a proxy to grammaticality of the questions for ranking and filtering.
\begin{enumerate}
    \item An \textbf{$n$-gram model} based on any pos-morph-tagged UD-compliant corpus. For instance, we use a 3-gram model, which could give the following probability $P(${\tt NOUN/Number=Sing}$|${\tt DET/Definite=Def}, {\tt ADJ/Number=Sing}$)$.
    \item A \textbf{question-word model}
    calculating the count $\mbox{\it c}(qw,r)$ for each pair of question word $qw$ and pos-morph expression of the root $r$ of the corresponding answer, e.g.,\ {\tt (when,NOUN/Number=Sing)}.

\end{enumerate}
Both models operate on \emph{pos-morph expressions} (i.e. words are substituted by their POS-tag together with UD morphological features, if applicable). For instance, the pos-morph expression for the word ``basketball'' is {\tt NOUN/Number=Sing} and for the word ``on'' is {\tt ADP}.

The first step is to filter out QA pairs with a single-word answer, whose pos-morph expression has never occurred in the training corpus as an answer. This step prevents generating single-word answers containing only function words (e.g., ``the'', ``himself'', ``to'').

The second step is to rank every remaining QA pair $j$ according to the score $r_{qa}^{(j)}$ using equations \eqref{eq:rank1} - \eqref{eq:rank3}. Equation \eqref{eq:rank1} is a convex combination using the n-gram model, where $p^*$ denotes the backoff probability, in which each word $w_i$ is substituted by its pos-morph expression (or by a POS-tag only if the former does not exist). $N_j$ is the number of trigrams in the question $q_j$, since we use a 3-gram model.
\begin{equation}
    r_{ng}^{(j)} = \frac{1}{N_j} \sum_{i=1}^{N_j}\lambda_1 p^*(w_{i}|w_{i-2}, w_{i-1}) + \lambda_2 p^*(w_{i}|w_{i-1}) + \lambda_3p^*(w_i) + \lambda_4\label{eq:rank1}
\end{equation}

Equation \eqref{eq:rank2} uses the question-word model to get the frequency of the pair of the question word $qw$ and the root token of the answer $r_{a_j}$.
\begin{equation}
    r_{qw}^{(j)} = \frac{\mbox{\it c}(qw,r_{a_j})}{\sum_i\mbox{\it c}(qw,r_i)}\label{eq:rank2}
\end{equation}

Equation \eqref{eq:rank3} provides the final score, which is a linear combination between the scores calculated in Equations \eqref{eq:rank1} and \eqref{eq:rank2}. $\alpha$ is a constant, which we set to 0.8 in our experiments.
\begin{equation}
    r_{qa}^{(j)} = \alpha \cdot r_{ng}^{(j)} + (1 - \alpha) \cdot r_{qw}^{(j)};\label{eq:rank3}
\end{equation}

Using the aforementioned $n$-gram and question-word models, a generated QA-pair is given a high score $r_{qa}$ if the question is made of a likely sequence of pos-morph expressions, and the question word (e.g., "when") matches the answer well.

The final step is referred to as \textbf{mean filtering}. This step ensures that only questions scoring higher than the mean of the scores for all generated questions for all sentences will be returned. Such filtering allows Quinductor to drop generated questions of potentially poor quality, and thus sometimes choose not to generate any QA-pairs for a given sentence.

\section{Data}
\label{sec:data}
To evaluate Quinductor in a multilingual setting we have utilized a dataset called TyDi QA \citep{clark2020tydi}. The dataset is a question-answering benchmark based on Wikipedia articles for 11 typologically diverse languages. 8 of these languages have available UD treebanks and trained dependency parsers in Stanza package \cite{qi2020stanza}, which we have utilized for inducing templates in all languages. For both training and evaluation we have excluded Yes/No-questions resulting in the training/development sets of the sizes reported in Table \ref{tab:tydiqa-stat}. Due to limited resources, we have performed a human evaluation only on a subset of languages, while reporting automatic evaluation metrics for all languages with the available UD treebanks.

To compare Quinductor to previous work we have also used the SQuAD dataset \citep{rajpurkar2016squad} for English, specifically the training/validation/test split provided by \citet{du2017learning}.

\begin{table}[t]
\caption{\label{tab:tydiqa-stat} Language-specific information along with the sizes of training and development splits (in the number of QA-pairs along with a proportion of the original training set) and the associated UD treebanks (UDT size, in tokens) used by the pre-trained Stanza parsers for the languages in the TyDi QA dataset. Question phrase positions are either obligatorily initial (OI), or not OI, or mixed, as defined by \citet{dryer200593}.}
\begin{tabular}{lccccc}
\midrule
\textbf{Language} & \textbf{QP position} & \textbf{Training set} & \textbf{Dev. set} & \textbf{UDT size} & \textbf{Human eval.} \\
\midrule
Finnish (fi) & OI & 7132 (47\%) & 1129 (52\%)& 397K & \greencheck\\
Russian (ru) & OI & 6425 (50\%)& 902 (56\%)& 1289K & \greencheck\\
English (en) & OI & 3837 (42\%)& 644 (62\%)& 648K & \greencheck\\
\midrule
Japanese (ja) & Not OI & 4506 (28\%) & 705 (41\%)& 1676K & \redcross\\
Telugu (te) & Not OI & 5680 (23\%)& 724 (29\%)& 6K & \redcross\\
Arabic (ar) & Not OI\tablefootnote{The exception is Syrian Arabic, in which the interrogative phrase is obligatorily initial.} & 14771 (64\%)& 1016 (74\%)& 1042K & \redcross\\
Indonesian (id) & Mixed & 5587 (37\%)& 728 (40\%)& 169K & \redcross\\
Korean (ko) & Not OI & 1638 (15\%)& 427 (25\%)& 446K & \redcross\\
\midrule
Bengali (bn) & Not OI & 2506 (23\%)& 129 (39\%)& NA & NA\\
Thai (th) & Not OI & 4150 (37\%)& 1161 (52\%)& NA & NA\\
Swahili (sw) & Not OI & 2372 (16\%)& 661 (29\%)& NA & NA\\
\bottomrule
\end{tabular}
\end{table}

\section{Evaluation}
\label{sec:evaluation}
Essentially, automatic evaluation metrics, such as BLEU \citep{papineni2002bleu}, ROUGE \citep{lin2004rouge}, METEOR \citep{agarwal2008meteor}, CIDEr \citep{vedantam2015cider}, rely on comparing word overlap between a generated question and a reference question. Such metrics can yield a low score even if the generated question is valid but just happens to be different from the reference question, or a high score even though the question is ungrammatical but happens to have a high word overlap with the reference question (see the article by \citet{callison2006re} for a further discussion). Nonetheless, \citet{amidei2018evaluation} report that 32\% of the surveyed papers on automatic QG used only automatic evaluation metrics, although \citet{nema2018towards} found that there is only a weak correlation between the automatic evaluation metrics and human judgements on answerability of the generated questions. For a broader discussion on the relationship between automatic evaluation metrics and human judgements an interested reader is referred to \citep[Section 7.4.1]{gatt2018survey}. In this article, we report automatic evaluation metrics for the following two reasons: Firstly, for the sake of comparability to other results reported in the literature, and giving a point of reference to researchers lacking resources to conduct human evaluations. Secondly, to assess the degree of the word overlap of the questions generated by Quinductor and the reference questions, as well as the quality of this overlap (e.g., if it contains mostly stop words).

The conducted human evaluation aims at providing insights about strengths and weaknesses of Quinductor as well as directions for future research. Unfortunately, there exist no standardized questionnaires and/or guidelines for human evaluation of automatically generated questions and answers. \citet{amidei2018evaluation} report 22 different criteria used by researchers to evaluate QG systems. Evaluations differ both on the number of criteria used and on the granularity of the rating scales for human judgements, see \citep[Table 7]{amidei2018evaluation} for more details. The number of human judges ranges from 1 to 364 (with an average of 4 and a mode of 2) and the number of sampled questions to be evaluated ranges from 60 to 2186 (with an average of 493). \citet{amidei2018evaluation} note that often the papers provide only little information about the evaluation guidelines as well.

For this article, we have tried to combine best practices from the reported evaluation guidelines for QG, notably \citep{heilman2009question, rus2010overview}, and more generally, best practices in human evaluation for NLG, as consolidated by \citet{van2020human}. On that basis, we propose to conduct human evaluation using a 9-item questionnaire. Each questionnaire item is rated on a 4-point Likert-type scale (see more information and design motivation in Appendix \ref{app:human_evaluation}).

A subset of automatic evaluation metrics (BLEU-N, ROUGE-L and CIDEr) were calculated using the nlg-eval package \citep{sharma2017nlgeval}, and METEOR using METEOR-1.5 package\footnote{METEOR-1.5 does not fully support all languages used in TyDi QA dataset, so we set language to ``other'' for all languages other than English)} \citep{denkowski2014meteor}. For all experiments we have used dependency parsers trained on UD treebanks as a part of Stanza package \citep{qi2020stanza}. All templates were induced and then processed using UDon2 \citep{kalpakchi2020udon2} -- an efficient package for manipulating dependency trees, written in C++ with Python bindings.

\subsection{Multilingual setting}
In order to support the claim about Quinductor's applicability to multiple languages, we have performed an evaluation on the TyDi QA dataset (see more information about the dataset in Section \ref{sec:data}). Different languages required somewhat different pre-processing steps, which are documented in Appendix \ref{app:data_preprocessing}.

\subsubsection{Automatic evaluation}
We have evaluated Quinductor on all languages present in the TyDi QA dataset with available UD treebanks and pre-trained dependency parsers in Stanza. The templates were induced using the training sets and the questions were generated on the development sets of the TyDi QA dataset. Only the top-ranked generated question (if any) was considered for automatic evaluation with the respective automatic evaluation metrics reported in Table \ref{tab:auto-eval}.
\begin{table}[t]
\centering
\caption{\label{tab:auto-eval} Automatic evaluation on the filtered TyDi QA development sets only for generated questions ranked first.}
\begin{tabular}{lcccccccc}
\midrule
\textbf{Metric} & \textbf{fi} & \textbf{ja} & \textbf{te} & \textbf{ar} & \textbf{id} & \textbf{ko} & \textbf{ru} & \textbf{en}\\
\midrule
BLEU-1 & 18.25 & 25.12 & 0 & 14.23 & 17.55 & 0 & 30.23 & 20.23\\
BLEU-2 & 10.04 & 12.03 & 0 & 8.35 & 10.06 & 0 & 19.99 & 12.16\\
BLEU-3 & 5.81 & 5.25 & 0 & 4.87 & 6.12 & 0 & 14.47 & 7.57\\
BLEU-4 & 3.42 & 2.30 & 0 & 2.90 & 3.74 & 0 & 11.23 & 4.72\\
METEOR & 11.75 & 12.03 & 0 & 13.12 & 11.67 & 0 & 19.02 & 12.46\\
ROUGE-L & 21.69 & 32.54 & 0 & 24.69 & 22.43 & 0 & 32.61 & 27.55\\
CIDEr & 7.0 & 22.29 & 0 & 22.70 & 26.51 & 0 & 63.69 & 21.35\\
\bottomrule
\end{tabular}
\end{table}

Quinductor was able to induce templates for all of these languages, but failed to generate any questions on the development sets for Telugu and Korean. The main reason is that QA-pairs for these languages contain answers that use smaller parts of some words (dubbed {\em subwords\/}) in the original sentence. As can be seen in Table \ref{tab:alg-prop}, such cases constitute 53\% and 60\% of the training QA-pairs for Telugu and Korean respectively, whereas the corresponding proportions for other languages are much lower. For instance, the word \begin{CJK}{UTF8}{mj}``이스라엘''\end{CJK} (``Israel'') is used as the answer for one of the QA-pairs in Korean, whereas the original sentence contains the word \begin{CJK}{UTF8}{mj}``이스라엘의''\end{CJK} (``Israeli''). Given that the number of possible questions not using subwords in the provided answers is only 19\%, and the dataset for Korean is the smallest (only 1638 QA-pairs), it is no surprise that Quinductor managed to generate only 9 templates. The same proportion for Telugu is 35\%, resulting in a larger number of templates, but only 1 generated question. This can most probably be attributed to the small size of Telugu's treebank (only 6K tokens), which might result in a less generalizable dependency parser.

\begin{table}[b]
\centering
\caption{\label{tab:alg-prop} Descriptive statistics of the TyDi QA training data for different languages, as well as templates and questions produced using this data. Recall that ``satisfactory questions'' have at least one word in common with the original sentence.}
\begin{tabular}{lcccccccc}
\midrule
\textbf{} & \textbf{fi} & \textbf{ja} & \textbf{te} & \textbf{ar} & \textbf{id} & \textbf{ko} & \textbf{ru} & \textbf{en}\\
\midrule
(1) Satisfactory questions & 74\% & 96\% & 68\% & 90\% & 83\% & 44\% & 62\% & 91\%\\
(2) Answer uses subwords & 9\% & 10\% & 53\% & 14\% & 5\% & 60\% & 10\% & 5\%\\
(1) but not (2) & 68\% & 86\% & 35\% & 78\% & 79\% & 19\% & 56\% & 87\%\\
Number of induced templates & 496 & 25 & 48 & 1104 & 340 & 9 & 85 & 254\\
Number of generated questions & 611 & 97 & 4 & 462 & 558 & 0 & 93 & 409\\
\bottomrule
\end{tabular}
\end{table}

While it is no surprise that subwords are used in agglutinative languages (e.g., Telugu, Korean, Japanese, Finnish, Indonesian) or fusional languages (e.g., Arabic), such cases are more surprising for English and Russian. For these languages, the cases are due to differences in tokenization between the original sentences and the provided answers (which can happen, since Stanza's tokenizers are based on neural networks). For example, the answer ``\$102 million'' was tokenized as ``\$102'', ``million'' for the answer and as ``\$'', ``102'', ``million'' in the original sentence. 

Russian and Japanese are two best performing languages in terms of BLEU-1 scores, meaning the induced questions have the highest word overlap with the reference questions. However, while Russian performs the best in terms of BLEU-4 (4-grams overlap), Japanese performs the worst (the other agglutinative languages, Finnish and Indonesian, perform similarly to Japanese in terms of BLEU-4).

Performance in METEOR scores (which have been shown by \citet{agarwal2008meteor} to correlate with human judgements better than BLEU scores) is roughly similar between all languages except a considerably higher score for Russian. This shows that while 4-gram precision is lower for some languages, the number of aligned matches is comparable.

Performance in ROUGE-L scores varies significantly with Japanese and Russian performing on-par at the top of the list, while Indonesian and Finnish are at the bottom of the list. While this clearly indicates that the length of the longest common matched subsequence varies across languages, the reasons behind this variation are unclear.

The final metric, CIDEr takes into account if the matched words are frequent or rare (and thus more informative) using inverse document frequency (IDF). The more rare the matched words, the higher the score. The CIDEr score for Russian is significantly higher than for all other languages meaning that word overlap with reference questions contains more rare words. By contrast, the score CIDEr for Finnish is significantly lower than for all the other languages, meaning that most of the matched words are frequent ones (such as, question words, prepositions or common verbs). This makes both Russian and Finnish interesting candidates for human evaluation to see whether such significant difference in CIDEr scores results in significant difference in the quality of the questions according to human judges.

The only available fusional language, Arabic, exhibits similar performance to the agglutinative languages (except CIDEr in Finnish). The notable difference is the significantly higher number of induced templates. It is also interesting that no templates could be generated without the prior removal of punctuation as a pre-processing step. This calls for additional investigation of the quality of the output of Arabic's dependency parser and potentially further tweaks of Quinductor to suit fusional languages better.

Finally, the performance for Indonesian is on-par with Arabic, which is interesting, given that Indonesian is the only language in TyDi QA with a mixed question phrase position (meaning that some question phrases are obligatorily initial and some are not). However, the templates for Indonesian have been induced assuming that the first word of the reference question is a question word. Hence the obtained performance might be due to specific properties of the dataset and requires further investigation on other datasets.

\subsubsection{Human evaluation}
As mentioned previously, Finnish and Russian were interesting candidates for human evaluation, and were chosen along with English. For evaluation, we randomly sampled 50 sentences for each language, and generated QA-pairs for them using the induced templates. 50 generated QA-pairs were combined with 50 original QA-pairs from the corpus (later referred to as {\em gold\/} QA-pairs), corresponding to the same sampled sentences, and presented for evaluation in a random order to 5 human judges via the Prolific platform\footnote{\url{https://www.prolific.co/}}. Each triple of a sentence and a QA-pair was judged using a questionnaire comprising 9 criteria (formulated as statements) to be evaluated on a 4-point Likert-type scale (from ``Disagree'' to ``Agree''). Further details about the questionnaires, guidelines and evaluation process in general are provided in Appendix \ref{app:human_evaluation}.

The score of each judge per criterion is treated as a judgement on an ordinal scale, instead of treating all criteria together as an interval scale. The rationale behind such treatment is that a single aggregated quality score of questions and/or answers (over judgement criteria) is not very informative and will not help in pinpointing the exact problems observed in generated QA-pairs.

Following the work of \citet{amidei2019agreement} we assess inter-annotator agreement (IAA) using Fleiss' $\kappa$ \cite{fleiss1971measuring} and Goodman-Kruskall's $\gamma$ \citep{goodman1979measures}. However, keeping in mind that we deal with ordinal data, the following two slight differences from \citep{amidei2019agreement} are introduced in our approach.

Firstly, Fleiss' $\kappa$ \citep{fleiss1971measuring} measures the level of agreement compared to agreement by chance, originally defined by Fleiss through the marginal distribution of scores over categories (hence another name of this statistics -- fixed-marginal $\kappa$). However, using Fleiss' $\kappa$ is appropriate only if judges know a priori how many cases should be distributed into each category (see \citep{randolph2005free} for an extensive discussion on the matter). In our case, it would not make sense to require judges to rate in this way, making the original Fleiss' $\kappa$ inappropriate for our purposes. Instead the free-marginal alternative $\kappa$, introduced by \citet{randolph2005free} and later referred to as Randolph's $\kappa$, should be used. In Randolph's $\kappa$ the probability of agreement by chance is assumed to be uniform and thus suitable in our case.

Secondly, Goodman-Kruskall's $\gamma$ (GK $\gamma$) was designed to measure rank correlation between ordinal judgements of two judges. \citet{amidei2019agreement} averaged GK $\gamma$ over pairs of judges, which is not interpretable from a statistical perspective, given that correlation coefficients are not additive (see Appendix \ref{app:gk_derivation} for a discussion on the matter). Instead of computing the mean, we propose a generalization of GK $\gamma$ to multiple raters, dubbed $\gamma_N$ (derived in Appendix \ref{app:gk_derivation}).
\begin{align}
    \Pi_N &= \{(i, j) | i \in U, j \in U, i < j \}\\
    C_N &= \sum_{(i, j) \in \Pi_N} C_{ij}; \ \ D_N = \sum_{(i, j) \in \Pi_N} D_{ij}; \ \ \gamma_N = \frac{C_N - D_N}{C_N + D_N}
\end{align}
where $U$ is the set of indices corresponding to human judges, $C_{ij}$ ($D_{ij}$) is the number of concordant pairs (i.e., ranked in the same order) or discordant pairs (ranked in the reversed order), between judges $i$ and $j$.

\begin{table}[t]
\caption{Inter-annotator agreement per criterion. Q stands for ``Question'' and SA -- for ``Suggested answer''}
\label{tab:human-eval-agreement}
\begin{tabular}{lccccccc}
\midrule
\multirow{2}{*}{\textbf{Criterion}} & \multirow{2}{*}{\textbf{IAA Metric}} & \multicolumn{2}{c}{\textbf{en}} & \multicolumn{2}{c}{\textbf{fi}} & \multicolumn{2}{c}{\textbf{ru}}\\
& & \textbf{gold} & \textbf{gen} & \textbf{gold} & \textbf{gen} & \textbf{gold} & \textbf{gen}\\
\midrule
\multirow{2}{*}{\parbox{4cm}{Q is grammatically correct}} & Randolph's $\kappa$ & 0.34 & 0.12 & 0.75 & 0.22 & 0.76 & 0.58\\
& GK $\gamma_N$ & 0.53 & 0.63 & 0.83 & 0.83 & 0.71 & 0.87\\
\midrule
\multirow{2}{*}{\parbox{4cm}{Q makes sense}} & Randolph's $\kappa$ & 0.25 & 0.17 & 0.67 & 0.29 & 0.76 & 0.61\\
& GK $\gamma_N$ & 0.55 & 0.72 & 0.78 & 0.82 & 0.79 & 0.93\\
\midrule
\multirow{2}{*}{\parbox{4cm}{Q would be clearer if more information were provided}} & Randolph's $\kappa$ & 0.15 & 0.09 & 0.49 & 0.35 & 0.40 & 0.42\\
& GK $\gamma_N$ & 0.44 & 0.47 & 0.53 & 0.62 & 0.46 & 0.56\\
\midrule
\multirow{2}{*}{\parbox{4cm}{Q would be clearer if less information were provided}} & Randolph's $\kappa$ & 0.41 & 0.36 & 0.85 & 0.78 & 0.59 & 0.81\\
& GK $\gamma_N$ & 0.47 & 0.54 & 0.86 & 0.88 & 0.65 & 0.93\\
\midrule
\multirow{2}{*}{\parbox{4cm}{Q is relevant to the given sentence}} & Randolph's $\kappa$ & 0.21 & 0.19 & 0.38 & 0.18 & 0.32 & 0.54\\
& GK $\gamma_N$ & 0.64 & 0.55 & 0.67 & 0.70 & 0.79 & 0.81\\
\midrule
\multirow{2}{*}{\parbox{4cm}{SA correctly answers the question}} & Randolph's $\kappa$ & 0.25 & 0.23 & 0.54 & 0.32 & 0.30 & 0.57\\
& GK $\gamma_N$ & 0.75 & 0.73 & 0.83 & 0.82 & 0.61 & 0.86\\
\midrule
\multirow{2}{*}{\parbox{4cm}{SA would be clearer if phrased differently}} & Randolph's $\kappa$ & 0.03 & 0.05 & 0.59 & 0.35 & 0.29 & 0.31\\
& GK $\gamma_N$ & 0.27 & 0.42 & 0.62 & 0.47 & 0.56 & 0.49\\
\midrule
\multirow{2}{*}{\parbox{4cm}{SA would be clearer if more information were provided}} & Randolph's $\kappa$ & 0.16 & 0.06 & 0.55 & 0.33 & 0.31 & 0.35\\
& GK $\gamma_N$ & 0.38 & 0.42 & 0.59 & 0.62 & 0.58 & 0.61\\
\midrule
\multirow{2}{*}{\parbox{4cm}{SA would be clearer if less information were provided}} & Randolph's $\kappa$ & 0.53 & 0.55 & 0.87 & 0.96 & 0.83 & 0.86\\
& GK $\gamma_N$ & 0.67 & 0.66 & 0.92 & 0.90 & 0.74 & 0.82\\
\bottomrule
\end{tabular}
\end{table}

Inter-annotator agreement (IAA) for the conducted human evaluations are reported in Table~\ref{tab:human-eval-agreement} per criterion for gold and generated QA-pairs separately. On the scale for Fleiss' kappa proposed by \citet{landis1977measurement}, there is a slight agreement between judges for most of the criteria for English, and a moderate agreement for Finnish and Russian. Following \citet{amidei2019agreement} we use the scale for GK $\gamma_N$ proposed by \citet{rosenthal1996qualitative}. On this scale, there is a large correlation between the judgements on most of the criteria for English, and a very large correlation for Finnish and Russian. A notable observation is that IAA for English is substantially lower on all criteria, no matter the IAA metric, or whether the QA-pairs were generated or gold.  Another observation of interest is that the generated QA-pairs get lower Randolph's $\kappa$, but higher GK $\gamma_N$ compared to the gold ones in the vast majority of cases. This means that the exact scores for generated questions differ more than for gold ones, but the ranking order is more consistent. 

\begin{figure}[H]
\centering
\includegraphics[width=0.72\textwidth]{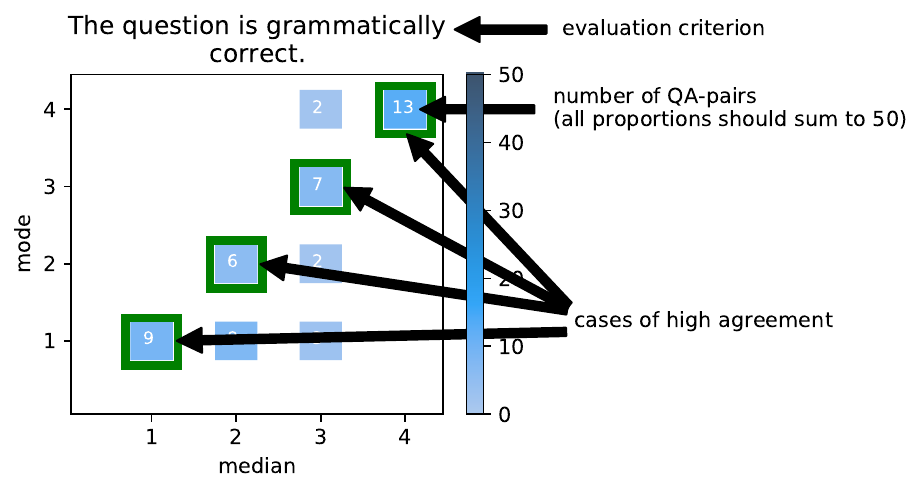}
\caption{An annotated example of a bi-variate histogram}
\label{fig:human_example_hist}
\end{figure}

\begin{figure}[hbtp]
\centering
\begin{minipage}{.45\textwidth}
	\centering
	\includegraphics[width=\textwidth]{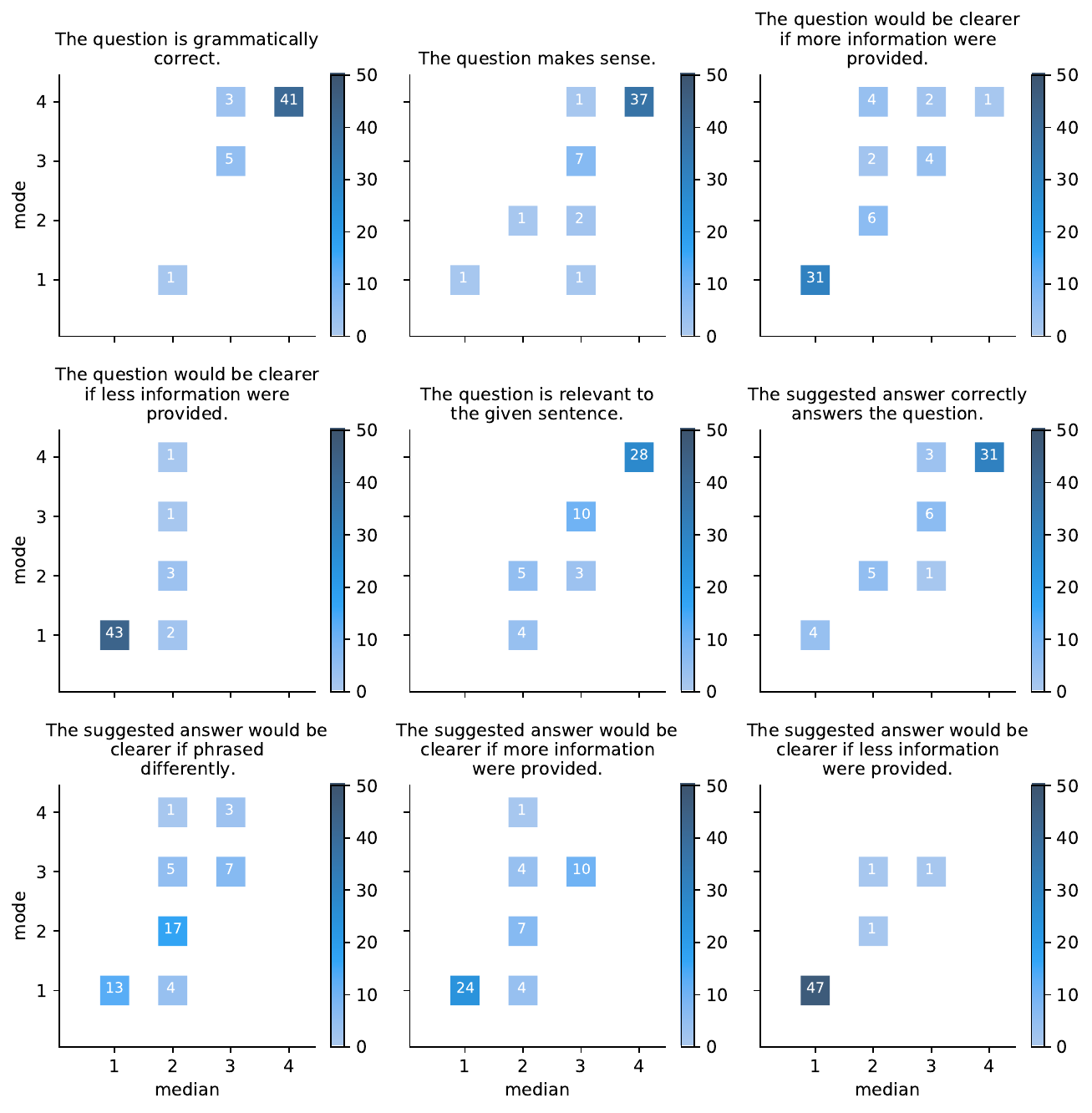}
	English (gold)
	\vspace{7px}
\end{minipage}
\begin{minipage}{.45\textwidth}
	\centering
	\includegraphics[width=\textwidth]{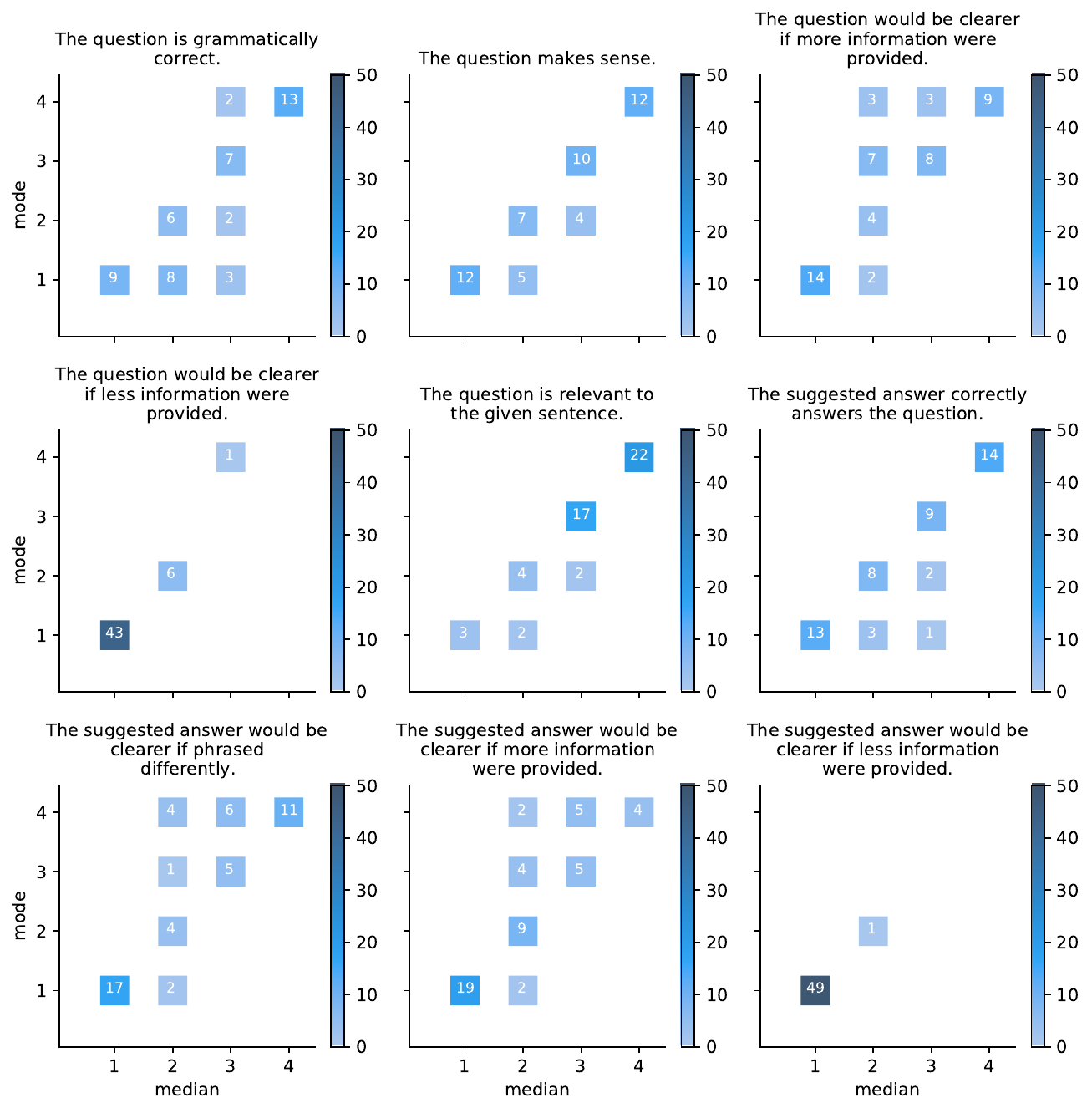}
	English (generated)
	\vspace{7px}
\end{minipage}
\begin{minipage}{.45\textwidth}
	\centering
	\includegraphics[width=\textwidth]{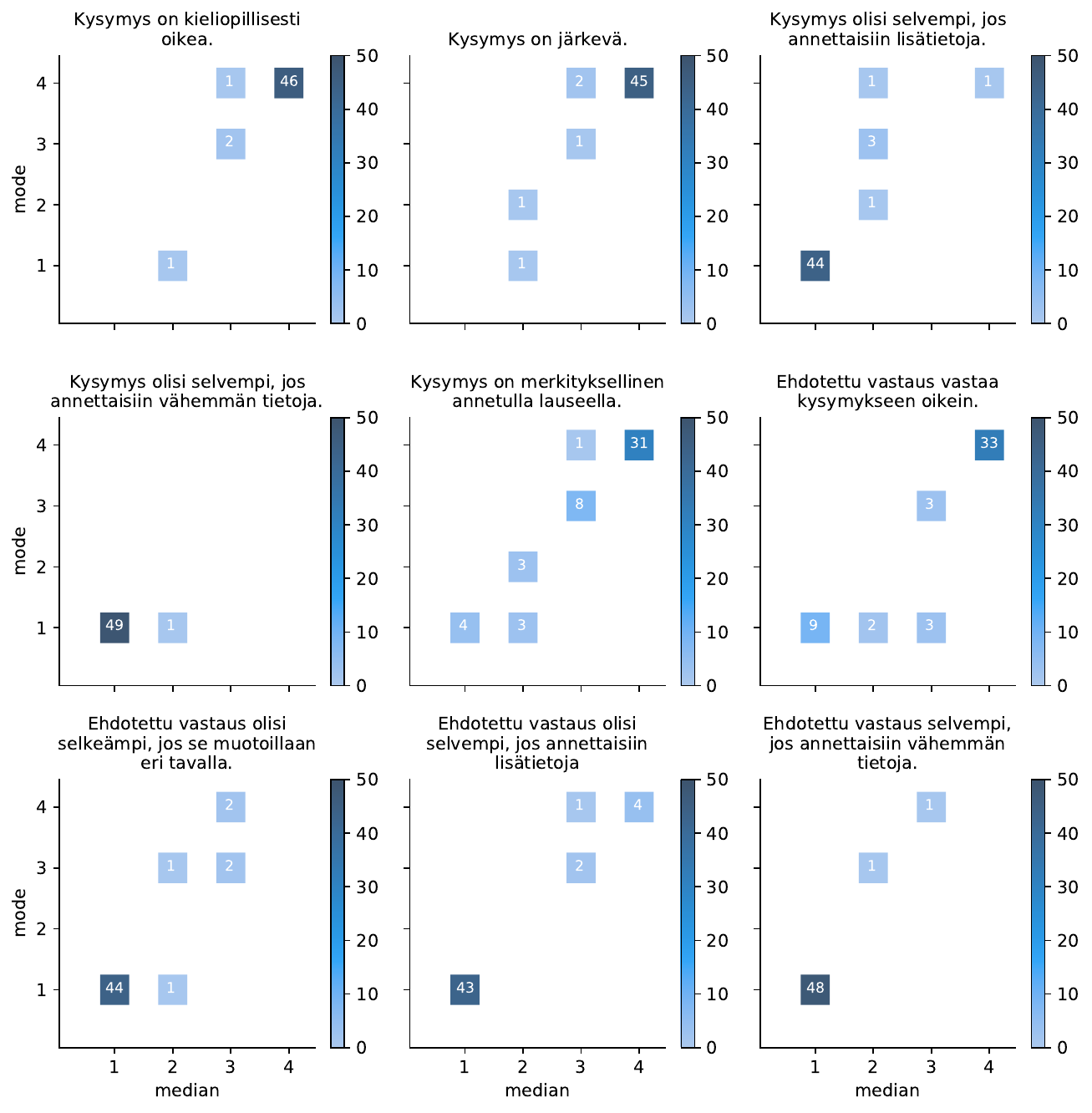}
	Finnish (gold)
	\vspace{7px}
\end{minipage}
\begin{minipage}{.45\textwidth}
	\centering
	\includegraphics[width=\textwidth]{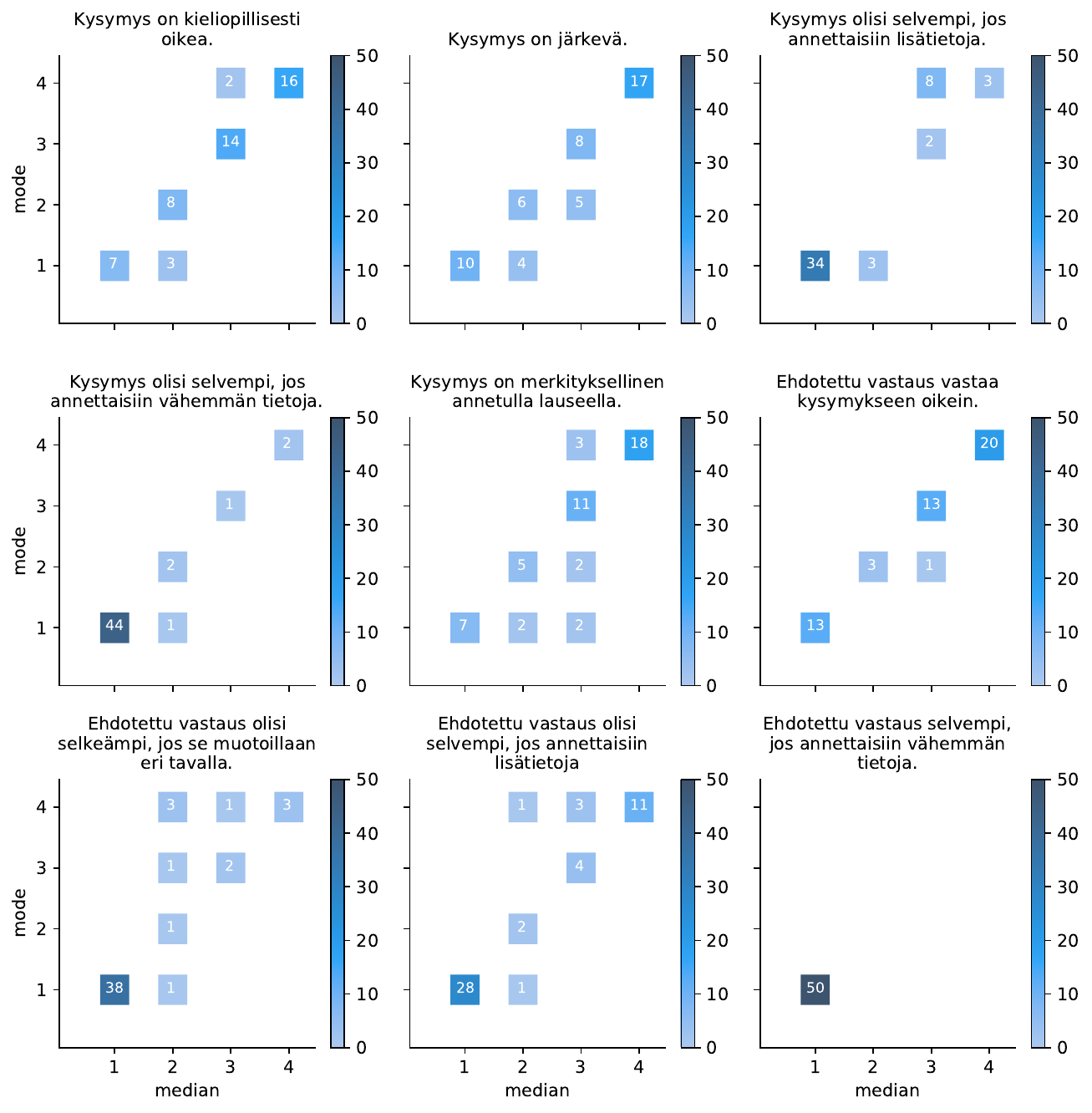}
	Finnish (generated)
	\vspace{7px}
\end{minipage}
\begin{minipage}{.45\textwidth}
	\centering
	\includegraphics[width=\textwidth]{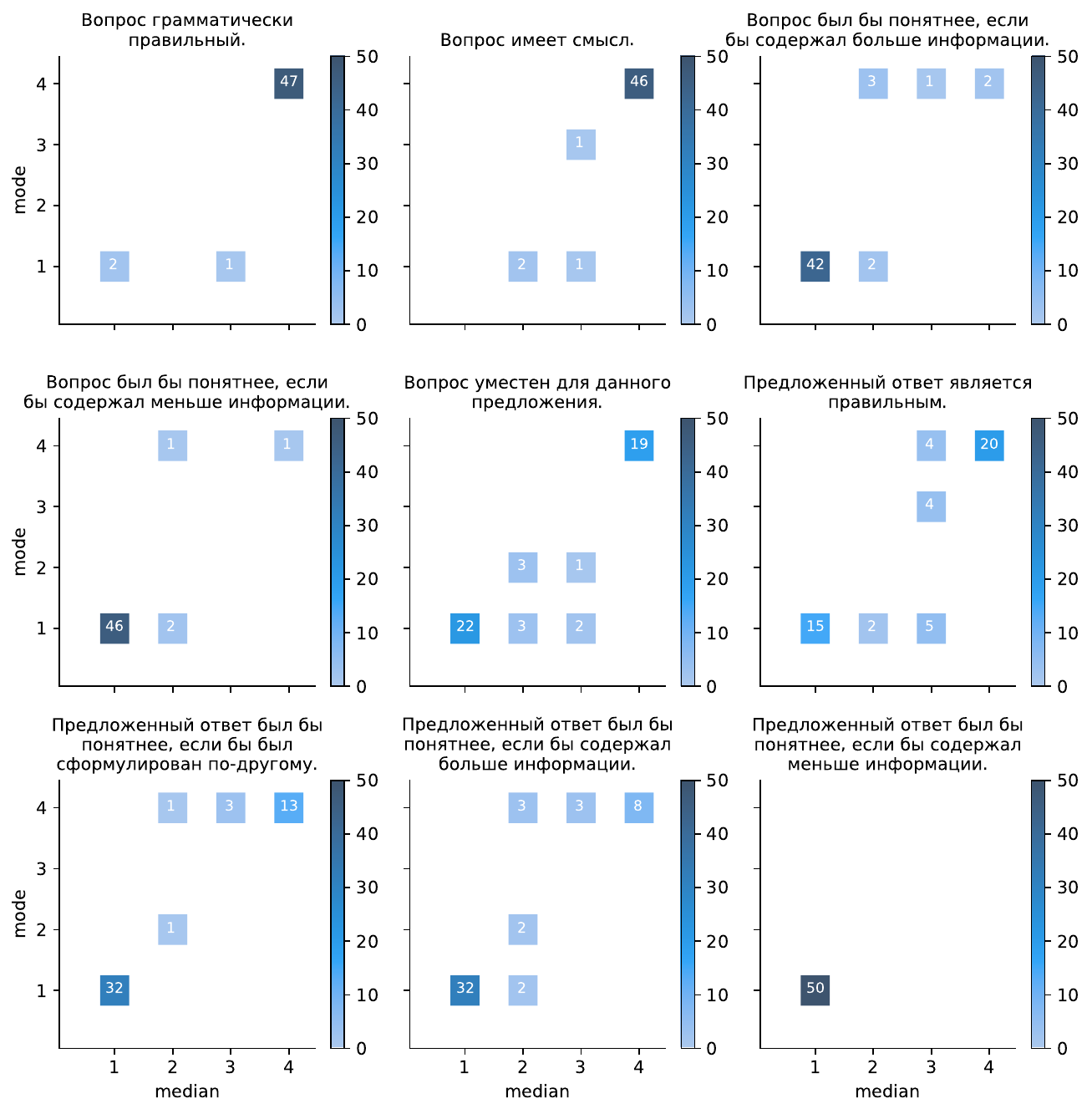}
	Russian (gold)
	\vspace{7px}
\end{minipage}
\begin{minipage}{.45\textwidth}
	\centering
	\includegraphics[width=\textwidth]{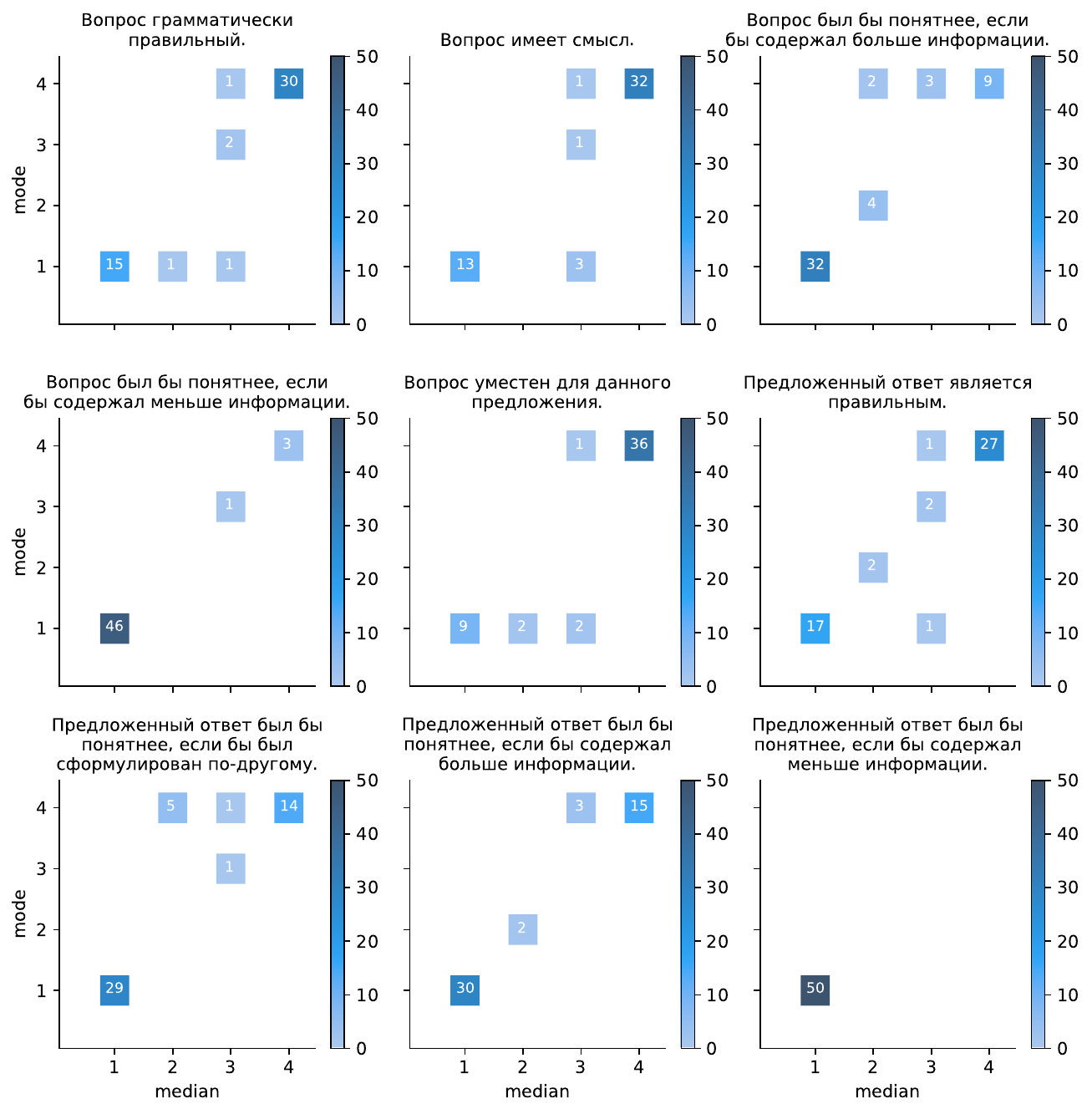}
	Russian (generated)
	\vspace{7px}
\end{minipage}
\caption{Bi-variate histograms of human judgements (the order of criteria is the same for all languages)}
\label{fig:human_fi_en_ru}
\end{figure}

To break down the results even further, we report the aggregated scores per each criterion using bi-variate histograms in Figure \ref{fig:human_fi_en_ru}. Recall that we treat human judgements as ordinal data. Valid measures of central tendency for ordinal data are {\em median\/} (the value separating a higher half from the lower half of a sample), and {\em mode\/} (the most frequent value of a sample), whereas {\em mean} is not a valid measure for ordinal scales (see \citep[Chapter~3]{blaikie2003analyzing} for more information). Hence the scores have been aggregated by median (on x-axis) and mode (on y-axis) over all 5 judges. If there are multiple modes, the worst one was taken, meaning if the ideal value for a criterion is ``Agree'' (corresponding to the numeric value of 4), then the smallest mode was considered, otherwise the largest. The rationale behind this handling of multi-modal distributions is to penalize cases where the human judges could not come to a definite agreement. To aid reader in understanding this presentation format, we provide an annotated histogram in Figure \ref{fig:human_example_hist}.

As established before, GK $\gamma_N$ is higher than Randolph's $\kappa$, meaning we should trust the exact ratings less and the relative rankings more. Then we are interested in QA-pairs ranked better than all other pairs by most of the judges, preferably that both median and mode for these QA-pairs are either equal to 4 if the best rating for the given criterion is 4, or to 1 if the best rating is 1. The proportion of such cases per criterion per language is presented in Table \ref{tab:human-eval-median-mode}. 

The majority of generated questions for English and Finnish are borderline (given 2 or 3) in terms of grammaticality, whereas the majority of questions in Russian were given a 4. It should be noted, though, that a considerable number of questions were evaluated as being grammatically incorrect (see Section \ref{sec:tydiqa_error_analysis} for error analysis). A similar pattern holds as to whether the question makes sense. The vast majority of the questions are not over-informative across all languages, and would not benefit from more information (except for English). Most of the cases for English and Finnish were borderline, whereas a substantial majority of questions were judged as relevant to the given sentence for Russian.

\begin{table}[t]
\centering
\caption{Proportion of generated QA pairs where both median and mode are the same}
\label{tab:human-eval-median-mode}
\begin{tabular}{lccccccc}
\midrule
\multirow{2}{*}{\textbf{Criterion}} & \multirow{2}{*}{\textbf{Best if}} & \multicolumn{2}{c}{\textbf{en}} & \multicolumn{2}{c}{\textbf{fi}} & \multicolumn{2}{c}{\textbf{ru}}\\
& & 1 & 4 & 1 & 4 & 1 & 4\\
\midrule
\parbox{4cm}{Q is grammatically correct} & 4 & 18\% & 26\% & 14\% & 32\% & 30\% & 60\%\\
\midrule
\parbox{4cm}{Q makes sense} & 4 & 24\% & 24\% & 20\% & 34\% & 26\% & 64\%\\
\midrule
\parbox{4cm}{Q would be clearer if more information were provided} & 1 & 28\% & 18\% & 68\% & 6\% & 64\% & 18\%\\
\midrule
\parbox{4cm}{Q would be clearer if less information were provided} & 1 & 86\% & 0\% & 88\% & 4\% & 91\% & 6\%\\
\midrule
\parbox{4cm}{Q is relevant to the given sentence} & 4 & 6\% & 44\% & 14\% & 36\% & 18\% & 72\%\\
\midrule
\parbox{4cm}{SA correctly answers the question} & 4 & 26\% & 28\% & 26\% & 40\% & 34\% & 54\%\\
\midrule
\parbox{4cm}{SA would be clearer if phrased differently} & 1 & 34\% & 22\% & 76\% & 6\% & 58\% & 28\%\\
\midrule
\parbox{4cm}{SA would be clearer if more information were provided} & 1 & 38\% & 8\% & 56\% & 22\% & 60\% & 30\%\\
\midrule
\parbox{4cm}{SA would be clearer if less information were provided} & 1 & 98\% & 0\% & 100\% & 0\% & 100\% & 0\%\\
\bottomrule
\end{tabular}
\end{table}

Focusing on the suggested answers, the majority of them have been reported to answer the question correctly for Finnish and Russian, whereas most of the cases were borderline for English. It should be noted that a substantial number of the suggested answers did not answer the question correctly. A breakdown of such cases is presented in Section \ref{sec:tydiqa_error_analysis}. A considerable number of answers would benefit from rephrasing for English and Russian, whereas the majority of answers for Finnish would not (which is surprising given that Finnish is an agglutinative language with rich inflectional morphology). Almost none of the suggested answers are over-informative, and a majority of them would not benefit from more information either (except for English).

\begin{table}[t]
\centering
\caption{\label{tab:error-grammar} Three most frequent types of grammatical mistakes for questions that received a mode of 1 or 2 for the criterion ``The question is grammatically correct''}
\begin{tabular}{lp{3.2cm}cp{6.7cm}}
\midrule
\textbf{Lang.} & \textbf{Problem} & \textbf{Freq.} & \textbf{Example} \\ 
\midrule
\multirow{4}{*}{en} & Wrong question word & 17.8\% & Who is the poorest state in the United States of America? \\
   & Underspecified & 17.8\% & Who finished career?\\
   & Wrong article & 14.3\% & Which is a largest hub?\\
\midrule
\multirow{6}{*}{ru} & A transitive verb lacks object & 47.1\% & \foreignlanguage{russian}{Когда архиепископ признал на ландтаге в городе?}\\
   & A split of a proper name & 11.8\% & \foreignlanguage{russian}{Когда Кеи» исключена компания «Мэри?}\\
   & Unresolved coreference & 11.8\% & \foreignlanguage{russian}{Когда состоялся третий шаг?}\\
\midrule
\multirow{5}{*}{fi} & Wrong question word & 61.1\% & Mik\"a oli genovalainen tutkimusmatkailija?\\
   & Question is nonsensical & 16.7\% & Milloin m\"a\"ar\"a olisi euroa?\\
   & Missing parts of question & 16.7\% & Min\"a vuonna ensimm\"ainen elokuva Spring of Birth sai?\\
\bottomrule
\end{tabular}
\end{table}

\subsubsection{Error analysis}
\label{sec:tydiqa_error_analysis}
Exploring the questions that obtained a mode of 1 or 2 for the grammaticality criterion, we have identified three most frequent types of errors, which are summarized in Table \ref{tab:error-grammar}. As can be seen, the types of errors are different across languages, suggesting that Quinductor's performance might be boosted for each individual language by applying language-specific post-processing (e.g., determiner correction for English). Some of the errors are connected to the errors in dependency parsing, such as split proper names in Russian, calling for a principled error analysis for these parsers beyond the provided development treebanks.

Another interesting issue pertains to questions that were judged grammatically correct (mode and median of 4 on the grammaticality criterion), but exhibited problems with respect to other criteria. Such cases are presented in Tables \ref{tab:error-right-grammar} and \ref{tab:error-right-grammar2}.

\begin{table}[b]
\centering
\caption{\label{tab:error-right-grammar} Examples of QA-pairs judged grammatically correct (median and mode of 4), but exhibiting problems in other criteria.}
\begin{tabular}{lcp{6.7cm}p{3.2cm}}
\midrule
\textbf{Lang.} & \textbf{ID} & \textbf{Question} & \textbf{Suggested answer}\\
\midrule
\multirow{2}{*}{en} & EN1 & Who served 18 months? & Susan McDougal  \\
   & EN2 & Where was the Nobel Peace Prize awarded? & Frédéric in 1901 \\
\hline
\multirow{4}{*}{ru} & RU1 & \foreignlanguage{russian}{Когда была основана компания?} & \foreignlanguage{russian}{тремя}\\
   & RU2 & \foreignlanguage{russian}{Когда скончался Эдуард?} & \foreignlanguage{russian}{5 января 1066}\\
   & RU3 & \foreignlanguage{russian}{Когда был закрыт монастырь?} & 1924\\
\hline
\multirow{4}{*}{fi} & FI1 & Minä vuonna erä päättyi? & 2.58\\
   & FI2 & Mitä sijamuodot ovat? & nominatiivi\\
   & FI3 & Mikä on tartuntatauti eli infektiotauti ( morbus contagiosus )? & infektiosairaus\\
\bottomrule
\end{tabular}
\end{table}

\begin{table}[t]
\centering
\caption{Human judgements of the examples in Table \ref{tab:error-right-grammar}. If only one number is specified, then mode and median are equal, otherwise the format is {\tt median/mode}}\label{tab:error-right-grammar2}
\begin{tabular}{lcccccccc}
\midrule
\textbf{Criterion} & \textbf{EN1} & \textbf{EN2} & \textbf{RU1} & \textbf{RU2} & \textbf{RU3} & \textbf{FI1} & \textbf{FI2} & \textbf{FI3}\\
\midrule
\parbox{4cm}{Q is grammatically correct} & 4 & 4 & 4 & 4 & 4 & 4 & 4 & 4\\
\midrule
\parbox{4cm}{Q makes sense} & 3 & 4 & 4 & 4 & 4 & 2/1 & 4 & 4\\
\midrule
\parbox{4cm}{Q would be clearer if more information were provided} & 4 & 4 & 1 & 4 & 2 & 2/1 & 1 & 1\\
\midrule
\parbox{4cm}{Q would be clearer if less information were provided} & 1 & 1 & 1 & 1 & 1 & 1 & 1 & 1\\
\midrule
\parbox{4cm}{Q is relevant to the given sentence} & 4 & 4 & 4 & 4 & 4 & 1 & 3/4 & 4\\
\midrule
\parbox{4cm}{SA correctly answers the question} & 4 & 1 & 1 & 4 & 2 & 1 & 1 & 3\\
\midrule
\parbox{4cm}{SA would be clearer if phrased differently} & 1 & 4 & 1 & 1 & 4 & 1 & 1 & 4\\
\midrule
\parbox{4cm}{SA would be clearer if more information were provided} & 2 & 4 & 1 & 1 & 4 & 1 & 4 & 4\\
\midrule
\parbox{4cm}{SA would be clearer if less information were provided} & 1 & 1 & 1 & 1 & 1 & 1 & 1 & 1\\
\bottomrule
\end{tabular}
\end{table}

As can be seen in Table \ref{tab:error-right-grammar2}, most of the questions also make sense, but would benefit from including more information. The suggested answers exhibit much more variation in human judgements, both in terms of them being correct, requiring rephrasing or more information. Most of the suggested answers are not over-informative, as also illustrated by the provided samples in Table \ref{tab:error-right-grammar2}.

\subsection{Comparison to other methods}
To support the claim of Quinductor being a strong baseline we compare our method to previously reported results for both state-of-the-art and baseline methods. Most of the previous work is done for the SQuAD dataset \citep{rajpurkar2016squad}, although the training/development/test split varies among articles, since the original SQuAD test set is hidden. We have found a number of articles relying on the SQuAD split\footnote{The SQuAD split is available at \url{https://github.com/xinyadu/nqg}} made by \citet{du2017learning} and others relying on the split made by \citet{zhou2017neural}. In this article use the former split and hence compare only to the articles that have explicitly reported to use of the same split to ensure a fair comparison between the methods. CIDEr is not provided in all other publications and is thus not reported. We induce templates based on the provided training set, and evaluate on the test set using automatic evaluation metrics only. The rationale for this is that some articles did not perform human evaluation at all \citep{kim2019improving,song2018leveraging,dong2019unified,zhao2018paragraph}, and others \citep{du2017learning,bahuleyan2017variational} used different criteria and evaluation guidelines making a fair comparison impossible.
\begin{table}[h]
\centering
\caption{\label{tab:eval-comp} Comparison to state-of-the-art QG methods and other reported baselines (shown in italics) on the test set of the SQuAD split made by \citet{du2017learning}}
\begin{tabular}{lccccccc}
\midrule
\textbf{Article} & \textbf{BLEU-1} & \textbf{BLEU-4} & \textbf{METEOR} & \textbf{ROUGE-L} \\
\midrule
\citep{dong2019unified} & NA & 22.12 & 25.06 & 51.07\\
\citep{kim2019improving} & NA & 16.2 & 19.92 & 43.96 \\
\citep{zhao2018paragraph} & 45.07 & 16.38 & 20.25 & 44.48\\
\citep{song2018leveraging} & NA & 13.98 & 18.77 & 42.72\\
\citep{du2017learning} & 43.09 & 12.28 & 16.62 & 39.75\\
\citep{bahuleyan2017variational} & 30.87 & 5.08 & NA & NA\\ 
\midrule
\emph{Vanilla seq2seq}\tablefootnote{The result fot this model is taken from \citep{du2017learning}\label{art3}} & 31.34 & 4.26 & 9.88 & 29.75\\
\emph{H\&S}\textsuperscript{\getrefnumber{art3}} & 38.50 & 11.18 & 15.95 & 30.98\\
\midrule
Ours & 30.56 & 9.71 & 16.70 & 31.71 \\
\bottomrule
\end{tabular}
\end{table}

As can be seen our method performs better than all reported baselines in terms of METEOR and ROUGE-L, and substantially better on BLEU-4 compared to the vanilla seq2seq model reported by \citet{du2017learning}.

\subsection{Cross-dataset evaluation}
In this final part of the evaluation, we explore how the induced templates for English are generalizing across datasets. We use 4889 templates induced from the SQuAD training set (from the split by \citet{du2017learning}), and 254 templates induced from the TyDi QA training set for English, to generate QA-pairs on the SQuAD test set (from the split by \citet{du2017learning}) and the TyDi QA development set. The results of this cross-dataset evaluation using automatic metrics are presented in Table \ref{tab:eval-squad-tydiqa}.

\begin{table}[h]
\centering
\caption{\label{tab:eval-squad-tydiqa} Automatic cross-dataset evaluation for first-ranked generated questions in English.}
\begin{tabular}{lccccc}
\midrule
\textbf{Training - test} & \textbf{BLEU-1} & \textbf{BLEU-4} & \textbf{METEOR} & \textbf{ROUGE-L} & \textbf{CIDEr} \\
\midrule
SQuAD - SQuAD & 30.56 & 9.71 & 16.70 & 31.71 & 7.69\\
SQuAD - TyDi QA & 13.47 & 2.22 & 10.79 & 23.09 & 11.33 \\
TyDi QA - TyDi QA & 20.23 & 4.72 & 12.46 & 27.55 & 21.35\\
TyDi QA - SQuAD & 34.31 & 11.12 & 14.83 & 30.61 & 8.84\\
\bottomrule
\end{tabular}
\end{table}

As can be observed, QA-pairs generated based on TyDi QA templates generally perform better than those based on SQuAD (except METEOR and ROUGE-L for the TyDi QA-SQuAD setup). This means that the word overlap is larger and of higher quality (i.e., consists of less common words) for the TyDi QA dataset. One reason for such performance difference is that mean filtering during the ranking step was designed for less-resourced languages, when only a few questions are generated. However, mean filtering is substantially weaker if many QA-pairs are generated, especially if most of them have low ranks (which is likely for SQuAD).

\section{Discussion}
\label{sec:discussion}
We have shown that the Quinductor method is a strong baseline method, outperforming baselines for English reported previously in the literature in terms of METEOR and ROUGE-L scores and performing better (or not far behind) some of the previously proposed QG methods. In addition, our method is inexpensive to train both in terms of time and textual resources, and thus applicable to languages other than English.

Quinductor has been successfully applied to 5 \emph{typologically diverse} less-resourced languages with limited training data. Most agglutinative languages (with a rich morphology and a free word order) performed similarly in terms of automatic evaluation metrics. Agglutinative languages with datasets relying on subwords in either questions or answers are proven to not work well with Quinductor (e.g., Korean and Telugu in TyDi QA dataset). Generated questions for Finnish performed better than English in terms of human judgements. Russian performed substantially better than all other languages both in terms of automatic evaluation metrics and human judgements, which might be a merit of a specific dataset and requires further investigation.

However, Quinductor has a number of limitations. Our method relies on the correctness of the dependency parser's output, or rather on the consistency of its errors. This assumption, although weaker than correctness, is still a limitation and does not always hold. We have noticed that some language-specific preprocessing techniques make the output of dependency parsers more consistent, but this requires further investigation.

Our method also incorporates a number of heuristics, such as mean filtering, selecting of contiguous template expressions in sentence transformation and ranking models, which seems to result in a comparable performance across languages. While only empirical evidence supports the applicability of these heuristics, we believe it is enough to make Quinductor a strong multilingual baseline, and set the lower bar for neural methods.

Another limiting property of Quinductor is that it lacks knowledge about semantics, since encoding such knowledge requires a large enough corpus that might not be available for all languages. While lack of semantic knowledge degrades the quality of questions, a surprisingly large number of them remain grammatically correct and make sense according to human evaluation.

It might be said that Quinductor still requires the use of a dependency parser to be trained on a sizeable dataset, and thereby moving the problem rather than solving it. However, firstly, the Universal Dependencies framework includes 200 treebanks for over 100 languages (and counting). Secondly, we have shown that Quinductor could induce templates even for Telugu, whose dependency parser is trained on a treebank with only 6K tokens. Thirdly, less-resourced languages have much smaller corpora of raw text, making pretraining of large-scale neural language models challenging (let alone fine-tuning them for QG). Finally, Quinductor method is a yet another use case for a dependency treebank, adding to the motivation of expanding UD to other languages.

\appendix
\begin{appendices}
\titleformat{\section}
{\normalfont\bfseries}
{\appendixname\ \thesection:}{0.5em}{}

\appsection{Human evaluation details}
\label{app:human_evaluation}
Human evaluation has been conducted on the Prolific platform\footnote{https://www.prolific.co/}. We used Prolific's pre-screening feature and required each human judge to have the language of interest as the first language and hold at least a high school diploma (A-levels). 

\begin{figure}[H]
	\centering
	\includegraphics[width=0.8\textwidth]{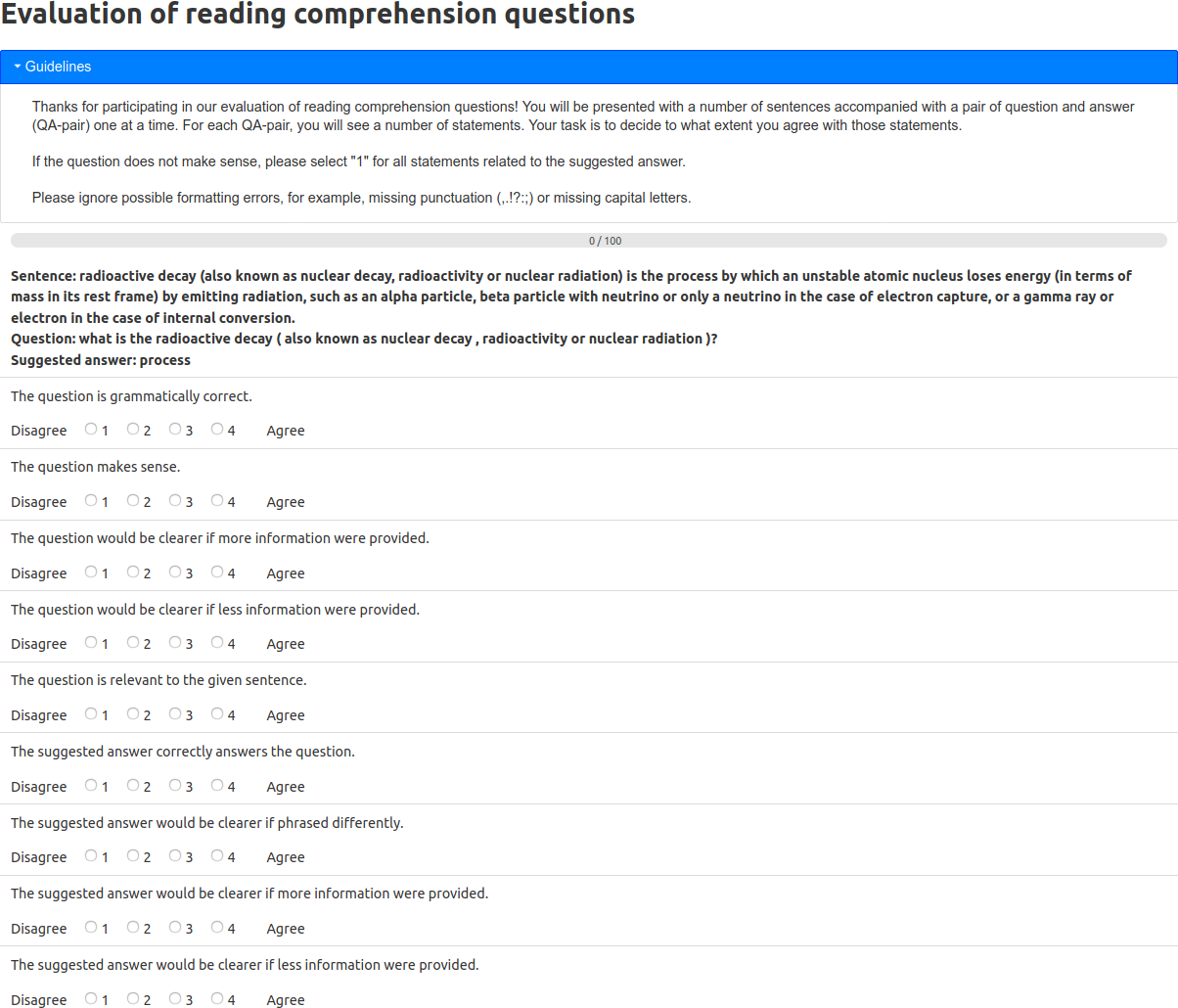}
	\caption{Evaluation guidelines and questionnaire for English}
	\label{fig:en_eval_guidelines}
\end{figure}

\begin{figure}[H]
	\centering
	\includegraphics[width=0.8\textwidth]{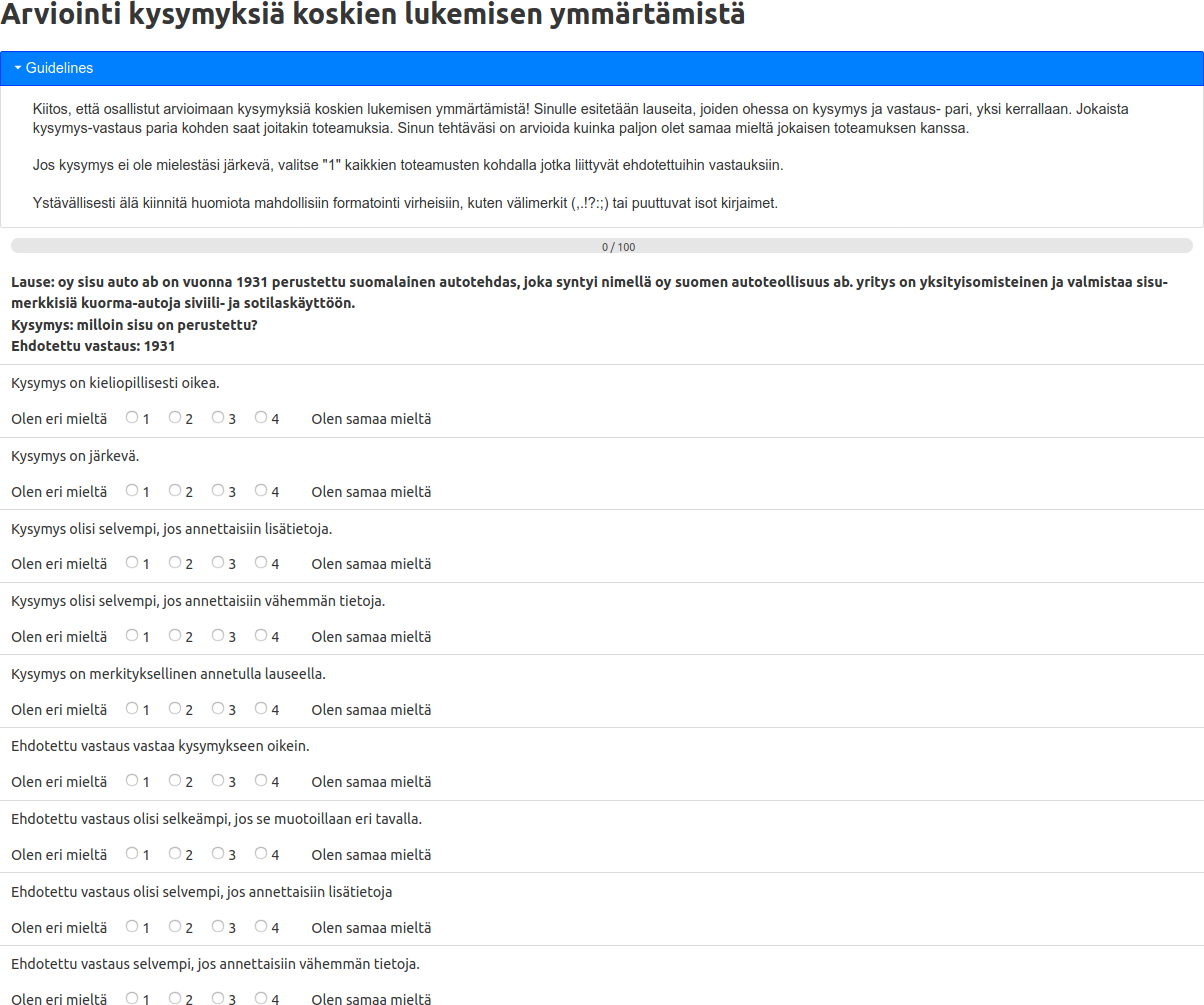}
	\caption{Evaluation guidelines and questionnaire for Finnish}
	\label{fig:fi_eval_guidelines}
\end{figure}

\begin{figure}[H]
	\centering
	\includegraphics[width=0.8\textwidth]{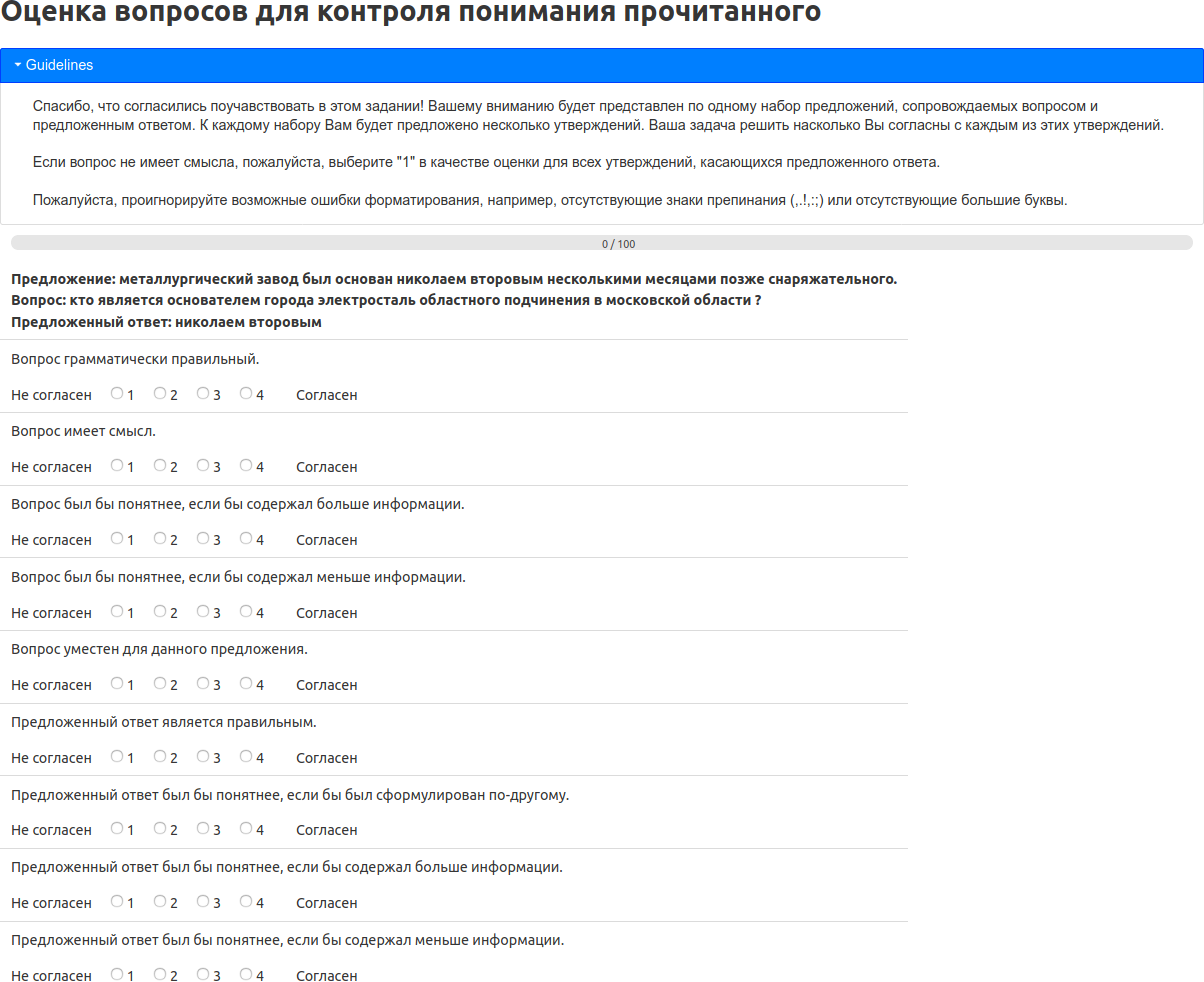}
	\caption{Evaluation guidelines and questionnaire for Russian}
	\label{fig:ru_eval_guidelines}
\end{figure}

The exact guidelines for human evaluation are presented in Figures \ref{fig:en_eval_guidelines}, \ref{fig:fi_eval_guidelines}, \ref{fig:ru_eval_guidelines}. The instructions and 9 evaluation criteria are the same, but are translated into every language. Each criterion is evaluated on a 4-point Likert-type scale with the ends labeled as ``Disagree'' and ``Agree''. A neutral option is excluded, to force judges to make a decision. We opted out of a more typical ``Strongly disagree'' -- ``Strongly agree'' scale to give judges some alternatives in the middle, such as, ``Somewhat (dis)agree''. Otherwise, the scale would be interpreted as ``Strongly disagree'' -- ``Disagree'' -- ``Agree'' -- ``Strongly agree'', which effectively collapses it to a binary scale.

\appsection{Data pre-processing steps}
The applied pre-processing steps for every language are specified in Table \ref{tab:preprocess}. Note that Arabic is written from right to left, whereas Quinductor processes sentences from left to right. Hence the assumed question word position for Arabic is the first processed word, which is effectively the end of the sentence for Arabic.
\label{app:data_preprocessing}
\begin{table}[H]
\centering
\caption{\label{tab:preprocess} Language-specific pre-processing steps before template induction. S denotes start of the sentence, E -- end of the sentence.}
\begin{tabular}{lcccccccc}
\midrule
\textbf{Preprocessing step} & \textbf{fi} & \textbf{ja} & \textbf{te} & \textbf{ar} & \textbf{id} & \textbf{ko} & \textbf{ru} & \textbf{en}\\
\midrule
Lowercase & \greencheck & NA & NA & NA & \greencheck & NA & \greencheck & \greencheck\\
Remove punctuation & \redcross & \greencheck & \greencheck & \greencheck & \redcross & \redcross & \redcross & \redcross\\
Remove diacritics & \redcross & \redcross & \redcross & \redcross & \redcross & \redcross & \greencheck & \redcross\\
Assumed question word position & S & E & E & S & S & E & S & S\\
\bottomrule
\end{tabular}
\end{table}

\appsection{Derivation of multi-rater Goodman-Kruskall's $\gamma_N$}\label{app:gk_derivation}
Let us start by presenting the original derivation of GK $\gamma$, proposed by \citet{goodman1979measures}, applied to the case of human evaluation. Assume we have two judges independently assigning scores from 1 to $\alpha$ to the same set of $M$ items. Let $s_{aj}$ denote a random variable associated with scores of judge $a$ to the item $j$. 
Let $c_{1,2}$ denote the event that a randomly selected pair of items will be ordered in the same way by judges $1$ and $2$ (such pair is called \textbf{concordant}). The probability of such event can then be calculated using Equation \eqref{eq:p_c}, given that the judgements are independent.
\begin{equation}\label{eq:p_c}
    P(c_{1,2}) = \sum_{(i,j) \in \Pi_M}P(s_{1i} < s_{1j})P(s_{2i} < s_{2j}) + P(s_{1i} > s_{1j})P(s_{2i} > s_{2j})
\end{equation}
Let $d_{1,2}$ denote the event that a randomly selected pair of items will be ordered differently by judges $1$ and $2$ (such pair is called \textbf{discordant}). The probability of such event can then be calculated using Equation \eqref{eq:p_d}, given that the judgements are independent.
\begin{equation}\label{eq:p_d}
    P(d_{1,2}) = \sum_{(i,j) \in \Pi_M}P(s_{1i} < s_{1j})P(s_{2i} > s_{2j}) + P(s_{1i} > s_{1j})P(s_{2i} < s_{2j})
\end{equation}
Let $t_{1,2}$ denote the event that a randomly selected pair of items will be get the same scores with each other (be \textbf{tied}) by both judges. Then the probability of ties is calculated using Equation \eqref{eq:p_t} given that the judgements are independent.
\begin{equation}\label{eq:p_t}
    P(t_{1,2}) = \sum_{(i,j) \in \Pi_M}P(s_{1i} = s_{1j}) + P(s_{2i} = s_{2j})
\end{equation}
In all equations above $\Pi_M$ is a 2-combination of the set of item indices between 1 and $M$.
The conditional probability of concordant orders given no ties ($\widetilde{t_{12}}$) then equals to:
\begin{equation}
    P(c_{1,2}|\widetilde{t_{1,2}}) = \frac{P(\widetilde{t_{1,2}}|c_{1,2}) P(c_{1,2})}{P(\widetilde{t_{1,2}})} = \frac{1 \cdot P(c_{1,2})}{1 - P(t_{1,2})} = \frac{P(c_{1,2})}{1 - P(t_{1,2})}
\end{equation}
Similarly, the conditional probability of discordant orders given no ties equals to $P(d_{1,2}|\widetilde{t_{1,2}}) = \frac{P(d_{1,2})}{1 - P(t_{1,2})}$. GK $\gamma$ was then proposed by \citet{goodman1979measures} to be computed as
\begin{equation}
    \gamma_{1,2} = \frac{P(c_{1,2}) - P(d_{1,2})}{1 - P(t_{1,2})}
\end{equation}
Observe that $P(c_{1,2}|\widetilde{t_{1,2}}) + P(d_{1,2}|\widetilde{t_{1,2}}) = 1$, since if there are no ties, there can be either concordant or discordant orders, then the following derivation holds:
\begin{align}
    P(c_{1,2}|\widetilde{t_{1,2}}) + P(d_{1,2}|\widetilde{t_{1,2}}) &= 1\\
    \frac{P(c_{1,2})}{1 - P(t_{1,2})} + \frac{P(d_{1,2})}{1 - P(t_{1,2})} &= 1\\
    \frac{P(c_{1,2}) + P(d_{1,2})}{1 - P(t_{1,2})} &= 1\\
    P(c_{1,2}) + P(d_{1,2}) &= 1 - P(t_{1,2})
\end{align}
Using this observation, GK $\gamma_{1,2}$ can be rewritten to a more familiar form:
\begin{equation}
    \gamma_{1,2} = \frac{P(c_{1,2}) - P(d_{1,2})}{P(c_{1,2}) + P(d_{1,2})}
\end{equation}

Now assume we have a third judge as well and we calculate $\gamma_{1,2}$, $\gamma_{1,3}$, and $\gamma{2,3}$. Then to evaluate the agreement between three judges we simply take an average of them. Let us see what it amounts to.
\begin{align}
    \gamma_{1-3} &= \frac{\gamma_{1,2} + \gamma_{1,3} + \gamma_{2,3}}{3}\\
    &= \frac{\frac{P(c_{1,2}) - P(d_{1,2})}{P(c_{1,2}) + P(d_{1,2})} + \frac{P(c_{1,3}) - P(d_{1,3})}{P(c_{1,3}) + P(d_{1,3})} + \frac{P(c_{2,3}) - P(d_{2,3})}{P(c_{2,3}) + P(d_{2,3})}}{3}\label{eq:gamma_final}
\end{align}
The quantity in Equation \ref{eq:gamma_final} cannot be simplified further resulting neither in a valid probability nor in a generalized version of GK $\gamma$.

Instead the derivation process can easily be extended to $N$ judges ($N > 2$), as follows, resulting in a generalized version of GK $\gamma$, dubbed $\gamma_N$. Let $c_N$, $d_N$ and $t_N$ be the events that a randomly selected pair of items is concordant, discordant or tied (respectively) by any pair selected from $N$ judges, then the following holds.
\begin{align}
    P(c_N) &= \sum_{(a,b) \in \Pi_N}\sum_{(i,j) \in \Pi_M}P(s_{ai} < s_{aj})P(s_{bi} < s_{bj}) + P(s_{ai} > s_{aj})P(s_{bi} > s_{bj})\\
    P(d_N) &= \sum_{(a,b) \in \Pi_N}\sum_{(i,j) \in \Pi_M}P(s_{ai} < s_{aj})P(s_{bi} > s_{bj}) + P(s_{ai} > s_{aj})P(s_{bi} < s_{bj})\\
    P(t_N) &= \sum_{(a,b) \in \Pi_N}\sum_{(i,j) \in \Pi_M}P(s_{ai} = s_{aj}) + P(s_{bi} = s_{bj})\\
    P(c_N|\widetilde{t_N}) &= \frac{P(c_N)}{1 - P(t_N)}\\
    P(d_N|\widetilde{t_N}) &= \frac{P(d_N)}{1 - P(t_N)}\\
    \gamma_N &= \frac{P(c_N) - P(d_N)}{P(c_N) + P(d_N)} = \frac{\sum\limits_{(a,b) \in \Pi_N} P(c_{a,b}) - \sum\limits_{(a,b) \in \Pi_N} P(d_{a,b})}{\sum\limits_{(a,b) \in \Pi_N} P(c_{a,b}) + \sum\limits_{(a,b) \in \Pi_N} P(d_{a,b})}
\end{align}
$\Pi_M$ is a 2-combination of the set of item indices between 1 and $M$, $\Pi_N$ is a 2-combination of the set of judge indices between 1 and $N$, 

\appsection{Samples of generated questions}
Here we present samples of generated questions for English, Finnish and Russian. All sentences, questions and suggested answers are lowercased, since we have empirically found Stanza's tokenizers and dependency parsers to perform more consistently when text is lowercased. Both QA-pairs and sentences were also lowercased for human judges during evaluation.

\qgsample{diphenhydramine was first made by george rieveschl and came into commercial use in 1946}{who made diphenhydramine?}{george rieveschl}
\qgsample{parallax (from ancient greek \foreignlanguage{greek}{παράλλαξις} (parallaxis), meaning 'alternation') is a displacement or difference in the apparent position of an object viewed along two different lines of sight, and is measured by the angle or semi-angle of inclination between those two lines.}{what is a displacement in the apparent position of an object viewed along two different lines of sight?}{parallax}
\qgsample{the lowest temperatures are registered in july and august (18°c - 64°f) and the highest in february (maximum temperature 28°c - 82.4°f [1]), the summer season in the southern hemisphere.}{where are the lowest temperatures registered?}{july}
\qgsample{the nobel peace prize was first awarded in 1901 to frédéric passy and henry dunant — who shared a prize of 150,782 swedish kronor (equal to 7,731,004 kronor in 2008) — and, most recently, to denis mukwege and nadia murad in 2018}{where was the nobel peace prize awarded?}{frédéric in 1901}
\qgsample{in 1986, the first statute aimed at defense contractor employee whistleblower protection was enacted.}{when was the first statute aimed at defense contractor employee whistleblower protection enacted?}{1986}

\qgsample{\foreignlanguage{russian}{кхл была образована в 2008 году и объединяла в себе первоначально 24 команды.}}{\foreignlanguage{russian}{когда была образована кхл?}}{\foreignlanguage{russian}{2008 году}}
\qgsample{\foreignlanguage{russian}{салли маргарет филд родилась в пасадине, калифорния, 6 ноября 1946 года в семье киноактрисы маргарет филд и армейского офицера ричарда драйдена[1].}}{\foreignlanguage{russian}{когда родилась салли маргарет филд?}}{6}
\qgsample{\foreignlanguage{russian}{14 июня 1952 в сша была заложена первая в мире апл «наутилус» (english: uss nautilus), и она была спущена на воду 21 января 1954 года[1][2][3].}}{\foreignlanguage{russian}{когда была заложена первая в мире апл «наутилус» ( english : uss nautilus )?}}{\foreignlanguage{russian}{14 июня 1952}}
\qgsample{\foreignlanguage{russian}{санатана родился в 1488 году в бенгальской деревне в провинции джессор.}}{\foreignlanguage{russian}{когда родился санатана?}}{1488}
\qgsample{\foreignlanguage{russian}{металлургический завод был основан николаем второвым несколькими месяцами позже снаряжательного}}{\foreignlanguage{russian}{когда был основан металлургическии завод?}}{\foreignlanguage{russian}{несколькими}}

\qgsample{jäämerentie oli valmistuessaan 531 kilometriä pitkä ja viisi metriä leveä.}{kuinka pitkä jäämerentie oli?}{531 kilometriä}
\qgsample{fennomania, suomenmielisyys, suomenkiihko[1] oli suomalaisten kansallisen heräämisen liike, joka syntyi 1800-luvun alkupuolella ja vaikutti erityisesti saman vuosisadan jälkipuolella.}{mitä fennomania oli?}{liike}
\qgsample{kaupungin väkiluku on noin 118000, ja sen pinta-ala on km², josta km² on vesistöjä.[1] kuopion keskustaajama sijaitsee kallaveteen etelästä työntyvällä kuopionniemellä, joka jakaa kallaveden kahteen toisistaan lähes erilliseen osaan.}{paljonko on kaupungin väkiluku?}{118000}
\qgsample{sen konsentraatio 25°c:n lämpötilassa on noin 1,004·10-7mol/l eli sen ph-arvo on 7,0}{paljonko on sen konsentraatio 25°c:n lämpötilassa?}{1,004·10-7mol/l}
\qgsample{pegaso oli ajoneuvojen tuotemerkki, joka kuului vuonna 1945 generalissimus francisco francon valtiollistamaa espanjan ajoneuvoteollisuutta yhdistämällä vuonna 1946 syntyneeseen, madridissa kotipaikkaansa pitäneeseen enasa:an [empresa nacional de autocamiones s.a.).}{mitä pegaso oli?}{tuotemerkki}

\end{appendices}
\section*{Acknowledgements}
\label{sec:acknowledgements}
This work was supported by Vinnova (Sweden's Innovation Agency) within  project 2019-02997. We would also like to thank Lisse-Lotte Hermansson for helping with translating instructions for human evaluation in Finnish, Kristiina Savola for help in assessing results of human evaluation for Finnish and Bram Willemsen for helpful comments and discussions on the matter of evaluation.

\bibliographystyle{compling}
\bibliography{compling}
\end{document}